\documentclass[a4paper,fleqn]{cas-sc}
\usepackage[numbers,sort&compress]{natbib}

\usepackage{booktabs,tabularx,array}
\usepackage{siunitx}
\usepackage{graphicx}
\usepackage{tikz}
\usetikzlibrary{arrows.meta,positioning,calc,fit,backgrounds,shapes.geometric}
\usepackage{algorithm}
\usepackage{algpseudocode}
\usepackage{subcaption}
\usepackage{placeins}
\usepackage{microtype}
\definecolor{ink}{HTML}{243B53}
\definecolor{accent}{HTML}{D97936}
\definecolor{teal}{HTML}{218C8D}
\definecolor{muted}{HTML}{75818D}
\definecolor{pale}{HTML}{F1F5F8}
\graphicspath{{figures/}}
\newcolumntype{Y}{>{\raggedright\arraybackslash}X}
\newcommand{\figref}[1]{Fig.~\ref{#1}}
\begin{document}
\let\WriteBookmarks\relax
\let\printorcid\relax
\shorttitle{A Dynamic Toolkit for Transmission Characteristics}
\shortauthors{Jiacheng Miao et al.}
\title[mode=title]{A Dynamic Toolkit for Transmission Characteristics of Precision Reducers with Explicit Contact Geometry}
\author[1,2]{Jiacheng Miao}
\cormark[1]
\author[1]{Chao Liu}
\author[1]{Qiliang Wang}
\author[3]{Yunhui Guan}
\author[2]{Weidong He}
\address[1]{Intelligent Equipment Division, CRRC Qishuyan Institute Co., Ltd, Changzhou, 213011, China}
\address[2]{School of Mechanical Engineering, Dalian Jiaotong University, Dalian, 116028, China}
\address[3]{Research Institute, CRRC Qishuyan Institute Co., Ltd, Changzhou, 213011, China}
\cortext[1]{Corresponding author. E-mail: haomjc@163.com}
\begin{abstract}
Transmission characteristics of precision reducers depend on the interaction
of contact geometry, bearing support, structural deformation, and loading
history. This paper presents a dynamic toolkit that connects distributed
local contacts to the complete mechanical reaction path. Work-conjugate
mappings transfer displacement, reaction, and tangent contributions between
contacts and rigid or reduced coordinates, with explicit allocation of
contact and body elasticity. An implicit generalized-$\alpha$ solution
distinguishes trial evaluations from accepted history. Contact-identified
records then supply performance protocols and configured geometric or
constitutive feedback.

The numerical demonstrations focus on mechanical coupling and pressure
recovery. An annular housing retains 156 interface components while its
structural coordinates are reduced. Two representations with 25 and 26
coordinates have similar static errors but cross-port response errors
of 24.975\% and 0.312\% over 10--5000~Hz relative to the same parent
model. In a shared-pin example, a prescribed $20~\mu$m radial displacement
of one wheel changes the load on a second, fixed wheel by approximately
75~N through cross-station compliance. Removing that compliance removes
this incremental transfer on the tested sleeve-seating branch.
A double-wheel cycloidal assembly relates branch-dependent torsional
response to internal contact-load variation and downstream pressure.
The pressure maxima remain resolution-sensitive despite small discrete
force residuals. Further performance and slow-state formulations specify
how the same motion and contact records support precision, vibration,
heat, wear, and durability models with their required physical inputs.
The framework separates the selection of model representation and
numerical resolution while retaining common definitions of motion,
force, and observation.
\end{abstract}
\maketitle
\section{Introduction}
\label{sec:introduction}

Precision reducers transmit torque through several interacting contacts.
Tooth profiles determine the available engagement, bearings determine how
the loaded bodies move, and the supporting structure redistributes their
reactions. A local change can consequently affect more than one transmission
characteristic: an altered clearance changes contact participation, the new
load distribution changes angular compliance, and the associated forces
excite the surrounding structure. For robotic joints, these relationships
must be evaluated over changes in load, speed, and direction. Recent
servomechanism experiments have likewise treated speed, load, and lubricant
temperature as joint inputs to transmission-error prediction and
compensation~\cite{g_bilancia2025_compensation}. A mechanical framework for
these questions must connect local geometry and deformation to the motion
and observations of the complete assembly.

Explicit contact modelling already provides an important part of this
connection. Contact-based multibody models resolve cycloidal wheels, pins,
and rolling-element bearings within one transmission
mechanism~\cite{g_xu2019_contact}. Three-dimensional semi-analytical
contact analysis includes profile and longitudinal modifications through
elastic influence coefficients~\cite{g_zhang2020_contact}, and a
quasi-static formulation combining clearance, deformation, and friction
predicts several characteristics, including accuracy, stiffness, load
capacity, and efficiency~\cite{g_qi2024_cdfm}. Recent design studies further
develop meshing-interval-based cycloidal
modification~\cite{g_gao2026_modification} and examine the combined effects
of profile modification and shaft-angle error in involute--cycloid internal
gearing~\cite{g_mo2026_modification}. These studies establish the value of
retaining geometric changes at the interface where they act.

The propagation of those changes depends on the surrounding load path.
Equivalent-mechanism formulations describe manufacturing and assembly
errors through closed kinematic loops~\cite{g_yang2021_error}.
The bi-directional drive transmission-error model of Wang et
al.~\cite{g_wang2024_bddte} relates transmission error and lost motion
while including the output mechanism in tolerance allocation.
Loaded contact analysis has also been incorporated into dynamic models
with profile modifications, manufacturing errors, and bearing
clearances~\cite{g_li2023_dynamic}. Under variable operation, angular
acceleration and load inertia contribute to transient transmission
error~\cite{g_xu2025_accuracy}. Thus the contact geometry, allowed body
motion, and applied loading must be specified together when interpreting
a precision prediction.

Structural flexibility adds coupling between contacts at different
locations. Recent torsional-stiffness analyses combine both reduction
stages, crankshafts, and bearings~\cite{g_li2025_stiffness}.
Flexible-multibody models with deformable cranks and wheels provide
another route, with finite-element and experimental
comparisons~\cite{g_li2025_flexible}. Spatial bearing--carrier models
extend the observation to moment rigidity and rotational
precision~\cite{g_xie2022_spatial}. Contact-dynamics studies of hysteresis
further include preload, clearance, and component
errors~\cite{g_xu2026_hysteresis}. Their interpretation requires a
measurement definition: loading rate and speed affect stiffness
tests~\cite{g_cheng2023_stiffness}, while nonlinear stiffness--friction
models support different lost-motion extraction
procedures~\cite{g_xu2025_lostmotion}. A local mesh stiffness and a measured
assembly stiffness therefore represent different stages of the mechanical
and observation problem.

Dynamic observations depend on how the internal forces reach the selected
measurement location. Reduced models of compound epicyclic--cycloidal
reducers describe modulation associated with time-varying contact
stiffness~\cite{g_tung2023_dynamic}. Bearing-defect and cycloidal-stage
fault models connect local interface changes to reducer
vibration~\cite{g_xu2024_bearing,g_e2025_fault}. Recent work distinguishes
modulation associated with meshing force from that introduced by the
transmission path~\cite{g_luo2025_modulation}; electromechanical models
extend the observation to motor-current
signatures~\cite{g_e2026_electromechanical}. These developments motivate
retaining both an identifiable excitation and the structural coordinates
through which its response is observed.

The necessary local resolution also depends on the question being asked.
Uniform arc-length discretization has been studied specifically for
cycloidal profiles in multibody contact
calculations~\cite{g_krol2025_discretization}. At a smaller physical scale,
transient tribo-dynamics resolves the interaction of pin motion,
lubricant-film squeezing, contact, and
wear~\cite{g_li2025_tribodynamics}. Increasing local detail, retaining
more structural modes, and changing the operating protocol serve different
purposes. They should be controlled independently, with each additional
model supplied with the loads, geometry, and time statistics that it
actually requires.

The literature thus provides coupled contact models, flexible assemblies,
and joint evaluations of several performance measures. General
flexible-multibody software also supplies reusable assembly and integration
methods~\cite{m_gerstmayr2024}. The methodological question addressed here is
how to retain the physical connections among these representations within
a reducer-specific framework. A contact load must remain associated with
its location, motion, and history; a structural reduction must preserve the
mapping at that contact; and a performance calculation must identify the
mechanical state and protocol from which its inputs were obtained. These
requirements become especially important when a model is reused with a
different structural representation or loading experiment.

Table~\ref{tab:method_comparison} places the present formulation alongside
closely related coupled models. The comparison concerns the physical
representation and the observations obtained from it; coupled contact,
flexible components, and joint performance prediction already have
established precedents.

\begin{table}[pos=tbp]
\caption{Reported representations and observation targets of selected
coupled reducer models. The comparison identifies methodological
relationships rather than ranking numerical accuracy or computational cost.}
\label{tab:method_comparison}
\centering\small
\setlength{\tabcolsep}{4pt}
\begin{tabularx}{\linewidth}{@{}>{\raggedright\arraybackslash}p{.20\linewidth}
>{\raggedright\arraybackslash}p{.37\linewidth}Y@{}}
\toprule
Study and system & Contact and structural representation &
Principal observations and methodological connection\\
\midrule
Xu et al. (2019)~\cite{g_xu2019_contact}\newline
Cycloidal RV stage &
Multibody contact dynamics with discretized cycloidal profiles,
multiple needle-roller contacts, and updates of the system generalized forces. &
Contact-load distribution and clearance-dependent internal load
transmission; a direct precedent for coupled gear--bearing reaction closure.\\
\addlinespace[3pt]
Qi et al. (2024)~\cite{g_qi2024_cdfm}\newline
Cycloidal drive &
Quasi-static force and torque balance with clearance and friction;
Hertz contact deformation and a fixed equivalent eccentric-shaft stiffness. &
Accuracy, backlash, stiffness, capacity, and efficiency within one
formulation; establishes the context for distinguishing the inputs and
protocols of multiple observations.\\
\addlinespace[3pt]
Li and Zhang (2025)~\cite{g_li2025_flexible}\newline
RV reducer &
Rigid--flexible multibody dynamics in ADAMS with flexible crankshafts
and cycloidal gears; contact and joint-based finite-element models
provide additional comparisons. &
Time-varying torsional stiffness, including test comparison;
connects elastic-component representation to the stiffness observed
under a specified operating condition.\\
\addlinespace[3pt]
Xu et al. (2026)~\cite{g_xu2026_hysteresis}\newline
RV reducer &
Contact dynamics with component geometric errors, bearing preload
and clearance, and torque-loading control based on a hysteresis test scheme. &
Hysteresis, torsional rigidity, and lost motion with experimental
comparison; links internal contact conditions to a defined loading-cycle observation.\\
\addlinespace[3pt]
Present work\newline
Reducer framework &
Local elastic laws, cross-contact compliance, and retained structural
coordinates with explicit elastic allocation and paired displacement,
reaction, and tangent maps. &
Structural transfer, shared-pin coupling, branch response, and pressure
resolution; a common evaluation-state convention connects local
calculations to mechanical solution and protocol-specific observation.\\
\bottomrule
\end{tabularx}
\end{table}

The present contribution makes the interfaces between these modelling
choices explicit: elasticity is allocated once, displacement and reaction
transformations also enter the tangent, and each observation retains its
mechanical evaluation state and loading protocol. The numerical studies
examine these connections at component and assembly levels under their
respective boundary conditions.

This paper presents a dynamic toolkit organized around those connections.
Here, \emph{explicit contact geometry} means direct evaluation of the
contacting profiles or surfaces. The mechanical equations are solved
implicitly using a Warp-based implementation. The contribution has three
connected parts, expressed through the elastic allocation, interface
mappings, and evaluation-state conventions of the reducer model:
\begin{enumerate}
 \item Displacement, reaction, and tangent maps connect distributed
 physical interfaces to the selected structural coordinates. Explicit
 allocation of local and body elasticity preserves cross-contact
 deformation while avoiding duplicate compliance contributions.
 \item Geometric force-arm terms and local algorithmic tangents enter
 the same assembly residual, with contact history held or advanced
 according to the trial and acceptance convention. This defines the
 connection between local nonlinear closure and implicit body motion.
 \item Identifiable contact loads, geometry, and exposure statistics
 pass from the mechanical solution to prescribed performance protocols
 and compatible slow-state updates. The pressure example demonstrates
 an independent local-resolution study tied to a fixed assembly state.
\end{enumerate}

The numerical studies examine these relationships at complementary levels.
An annular structure demonstrates the consequences of structural reduction
at a fixed contact interface. A shared output pin isolates load transfer
between two wheel contacts. A double-wheel cycloidal assembly then connects
the external torque response to aggregate contact-load variation and
normalized downstream pressure. Component checks and the local resolution study
identify the scope of the quantitative evidence. Together, the formulation
and examples establish how the framework preserves load-path interactions
and specifies the independent idealized examples. The reducer application
withholds its detailed geometric inputs; its published results are
restricted to the stated aggregate and normalized observations.

\section{A common model of the transmission and its load paths}
\label{sec:framework}

The framework represents a reducer as an assembly in which motion,
contact load, and structural deformation determine one another.
Tooth geometry establishes where force can be transmitted, but the
operating separation and alignment also depend on the shafts, bearings,
output connections, and installation. Their reactions depend in turn
on the tooth forces. Closing this relation requires the local contact
problem to retain its connection to the surrounding assembly.
The organization adopted here follows contact-based multibody
modelling~\cite{m_gerstmayr2024} and develops the interfaces needed to
represent these reducer-specific interactions.

The information retained at an interface includes its physical location,
relative motion, normal and tangential force, and applicable internal
history. Structural models receive these loads and return displacements
at the corresponding sites; the assembly converts local forces into
component forces and moments. This association between a quantity, its
location, and its state is central to the framework. It allows an output
response to be examined together with the local load redistribution
that produced it, and allows local stress calculations to use the
appropriate operating contact rather than an inferred mean load.

\subsection{Transmission topology and reaction-force closure}

Transmission topology specifies how the bodies and interfaces are
connected. In a single-stage cycloidal arrangement, an eccentric input
drives one or two wheels and transfers their motion through pin--hole
output contacts. RV arrangements introduce preliminary power-splitting
gears and multiple cranks, with an additional offset input and compound
gear in the three-stage form. These arrangements share a downstream
structure of main wheels, eccentric bearings, crank supports, and an
output carrier. Cycloidal pin gearing and involute internal gearing
with a small tooth-number difference provide alternative main-stage
geometries. A change of gear type also requires consistent profiles,
eccentricity, assembly phase, inertia, and local constitutive data.
An implemented hypoid-contact extension uses its own paired-surface
and shaft assumptions within the same force-and-motion organization.

Figure~\ref{fig:load_paths} identifies the mechanical connections that
close the reaction path. In the supported RV configurations, crank
translations and the selected axial and tilt coordinates are determined
through bearing forces. Opposed bearing rows balance axial loads and
moments according to their axial spacing, contact orientations, and
assembly closure. The load passes between the main contacts,
eccentric journals, crank supports, carrier, main bearings, and
installation. In the double-wheel cycloidal arrangement, a common
output pin instead receives the two wheel loads at separate axial
stations. Fixed-axis constraints remain available where the model
deliberately imposes an alignment. The choice of retained coordinates
therefore specifies which alignment changes can emerge from loading.

\begin{figure}[pos=tbp]
\centering
\resizebox{\linewidth}{!}{\begin{tikzpicture}[x=1cm,y=1cm,
  font=\sffamily\fontsize{8}{9.5}\selectfont,
  line cap=round,line join=round,
  metal/.style={draw=ink,line width=.7pt,fill=pale},
  wheel/.style={draw=teal!85!black,line width=.8pt,fill=teal!10},
  shaft/.style={draw=ink,line width=.75pt,fill=ink!12},
  motion/.style={draw=accent,line width=1.05pt,-{Stealth[length=1.8mm]}},
  exchange/.style={draw=teal,line width=.8pt,{Stealth[length=1.4mm]}-{Stealth[length=1.4mm]}},
  leader/.style={draw=muted,line width=.5pt},
  fine/.style={font=\sffamily\fontsize{7}{8.2}\selectfont,text=muted},
  title/.style={font=\sffamily\bfseries\fontsize{9}{10.5}\selectfont,text=ink},
  pics/bearing/.style={code={
    \draw[metal](-.19,-.19)rectangle(.19,.19);
    \draw[ink,line width=.5pt](-.13,-.13)--(.13,.13)
                            (-.13,.13)--(.13,-.13);}},
  pics/main/.style={code={
    \draw[metal](-.22,-.16)rectangle(.22,.16);
    \draw[ink,line width=.5pt](0,0)circle(.105);}}]
\path[use as bounding box](0,-.3)rectangle(16,12.3);

\begin{scope}[yshift=6.35cm]
\node[title,anchor=west]at(.05,5.6){(a) Single stage: two wheels share each output pin};

\draw[ink,line width=1.1pt](1.55,4.65)--(9.55,4.65);
\foreach \x in {1.7,2.0,...,9.5}{\draw[leader](\x,4.65)--++(.12,.14);}
\node[anchor=south]at(5.55,4.97){Pin ring / housing};
\draw[metal](2.02,1.22)rectangle(2.38,4.29);
\draw[metal](8.72,1.22)rectangle(9.08,4.29);
\draw[metal](2.02,1.04)rectangle(9.08,1.30);
\node[fill=white,inner sep=2pt]at(5.55,1.17){Shared output flange / carrier};
\draw[motion](9.13,1.18)--(10.45,1.18);
\node[anchor=south,text=accent]at(9.78,1.37){$T_{\rm out}$};
\pic at(2.20,4.46){main}; \pic at(8.90,4.46){main};
\node[anchor=east]at(1.82,4.45){$M_L$};
\node[anchor=west]at(9.26,4.45){$M_R$};
\draw[exchange](2.66,4.10)--(2.66,4.61);
\draw[exchange](8.44,4.10)--(8.44,4.61);

\foreach \x/\w in {4.2/0,6.7/1}{
  \draw[wheel](\x-.30,1.95)rectangle(\x+.30,4.01);
  \node[text=teal!75!black,fill=teal!10,inner sep=1pt]at(\x,3.83){$W_{\w}$};
  \fill[white](\x-.29,3.20)rectangle(\x+.29,3.61);
  \fill[white](\x-.29,2.22)rectangle(\x+.29,2.80);
}
\draw[shaft](3.55,4.04)rectangle(7.35,4.17);
\draw[ink,line width=.85pt](5.45,4.17)--(5.45,4.65);
\draw[exchange](4.20,4.02)--(4.20,4.60);
\draw[exchange](6.70,4.02)--(6.70,4.60);

\draw[shaft](2.20,3.30)rectangle(8.90,3.51);
\draw[draw=accent,line width=.85pt,fill=accent!14]
  (3.57,3.23)rectangle(7.33,3.59);
\draw[ink,line width=.65pt](3.57,3.33)--(7.33,3.33)
                          (3.57,3.49)--(7.33,3.49);
\draw[exchange](4.69,3.04)--(4.69,3.39);
\draw[exchange](6.21,3.04)--(6.21,3.39);
\node[anchor=south]at(5.45,3.64){Sleeve};
\draw[leader](7.72,3.38)--(8.10,3.76)--(9.25,3.76);
\node[anchor=west,fine]at(9.28,3.76){Pin support};

\draw[shaft](.55,2.43)rectangle(9.35,2.59);
\draw[shaft](3.86,2.43)rectangle(4.54,2.76);
\draw[shaft](6.36,2.25)rectangle(7.04,2.59);
\pic at(4.20,2.65){bearing};\pic at(6.70,2.36){bearing};
\pic at(2.20,2.51){bearing};\pic at(8.90,2.51){bearing};
\node[anchor=east]at(1.82,2.10){$S_L$};
\node[anchor=west]at(9.26,2.10){$S_R$};
\node[anchor=north]at(4.20,1.91){$E_0$};
\node[anchor=north]at(6.70,1.91){$E_1$};
\draw[motion](.15,2.51)--(1.30,2.51);
\node[anchor=south,text=accent]at(.80,2.79){Input};
\draw[exchange](2.20,2.03)--(2.20,1.55);
\draw[exchange](8.90,2.03)--(8.90,1.55);
\node[fine,anchor=north]at(5.47,.88)
 {Axial stations: $z_{S,L}$, $z_{W,0}$, $z_{W,1}$, $z_{S,R}$};

\begin{scope}[shift={(13.6,3.3)}]
  \draw[ink,line width=.85pt](0,0)circle(1.25);
  \draw[wheel](0,.09)circle(1.05);
  \draw[muted,densely dashed](0,0)circle(.72);
  \foreach \a in {0,60,...,300}{
    \draw[draw=accent,fill=accent!16,line width=.7pt]
       (\a:.72)circle(.125);}
  \draw[shaft](0,0)circle(.24);
  \draw[teal,line width=.65pt](0,.09)circle(.34);
  \node[font=\sffamily\fontsize{7}{8}\selectfont]at(0,0){$C$};
\end{scope}
\node[anchor=south,align=center]at(13.6,4.85){Transverse topology};
\node[fine,align=center]at(13.6,1.58)
 {Central input $C$; output-pin array\\One pin is sectioned at left};
\node[fine,anchor=west]at(.05,.25)
 {Supported configuration: paired shaft and main bearings; one common pin shown.};
\end{scope}

\draw[muted!40,line width=.5pt](.05,6.27)--(15.95,6.27);

\begin{scope}
\node[title,anchor=west]at(.05,5.90){(b) RV stages: several cranks share the wheels and carrier};
\draw[ink,line width=1.1pt](1.55,4.65)--(9.55,4.65);
\foreach \x in {1.7,2.0,...,9.5}{\draw[leader](\x,4.65)--++(.12,.14);}
\node[anchor=south]at(5.55,4.97){Fixed pin ring or internal ring gear / housing};
\draw[metal](2.02,1.22)rectangle(2.38,4.29);
\draw[metal](8.72,1.22)rectangle(9.08,4.29);
\draw[metal](2.02,1.04)rectangle(9.08,1.30);
\node[fill=white,inner sep=2pt]at(5.55,1.17){Shared output carrier};
\draw[motion](9.13,1.18)--(10.45,1.18);
\node[anchor=south,text=accent]at(9.78,1.37){$T_{\rm out}$};
\pic at(2.20,4.46){main};\pic at(8.90,4.46){main};
\node[anchor=east]at(1.82,4.45){$M_L$};
\node[anchor=west]at(9.26,4.45){$M_R$};
\draw[exchange](2.66,4.10)--(2.66,4.61);
\draw[exchange](8.44,4.10)--(8.44,4.61);
\foreach \x/\w in {4.2/0,6.7/1}{
  \draw[wheel](\x-.30,1.95)rectangle(\x+.30,4.01);
  \node[text=teal!75!black,fill=teal!10,inner sep=1pt]at(\x,3.79){$W_{\w}$};
  \fill[white](\x-.29,2.36)rectangle(\x+.29,3.21);
}
\draw[shaft](3.55,4.04)rectangle(7.35,4.17);
\draw[ink,line width=.85pt](5.45,4.17)--(5.45,4.65);
\draw[exchange](4.20,4.02)--(4.20,4.60);
\draw[exchange](6.70,4.02)--(6.70,4.60);

\draw[shaft](.78,2.69)rectangle(9.35,2.85);
\draw[shaft](.94,2.32)rectangle(1.18,3.24);
\foreach \y in {2.38,2.54,...,3.18}{\draw[ink,line width=.5pt](.90,\y)--(1.22,\y);}
\draw[shaft](3.86,2.69)rectangle(4.54,3.12);
\draw[shaft](6.36,2.42)rectangle(7.04,2.85);
\pic at(4.20,3.00){bearing};\pic at(6.70,2.54){bearing};
\pic at(2.20,2.77){bearing};\pic at(8.90,2.77){bearing};
\node[anchor=east]at(1.82,1.96){$S_{m,L}$};
\node[anchor=west]at(9.26,1.96){$S_{m,R}$};
\node[anchor=north]at(4.20,1.91){$E_{m,0}$};
\node[anchor=north]at(6.70,1.91){$E_{m,1}$};
\node[anchor=south,fill=white,inner sep=1pt]at(5.55,3.31){Floating crank $C_m$};
\draw[exchange](2.20,2.07)--(2.20,1.55);
\draw[exchange](8.90,2.07)--(8.90,1.55);
\draw[motion](.22,3.82)--(1.05,3.82)--(1.05,3.30);
\node[align=center,fine,text=accent]at(.66,4.19)
 {Gear\\drive};
\node[fine,anchor=north]at(5.47,.88)
 {One offset crank section; the same connections repeat for $m=1,\ldots,N_c$};

\begin{scope}[shift={(13.6,3.3)}]
  \draw[ink,line width=.85pt](0,0)circle(1.25);
  \draw[wheel](0,.09)circle(1.05);
  \draw[muted,densely dashed](0,0)circle(.69);
  \foreach \a/\m in {90/1,210/2,330/3}{
    \draw[shaft](\a:.69)circle(.22);
    \node[font=\sffamily\fontsize{7}{8}\selectfont]at(\a:.69){$C_{\m}$};}
  \draw[muted](0,0)circle(.17);
\end{scope}
\node[anchor=south,align=center]at(13.6,4.85){Transverse topology};
\node[fine,align=center]at(13.6,1.58)
 {Cranks lie on a common pitch circle\\Shared wheels and carrier couple the branches};
\node[fine,anchor=west]at(.05,.25)
 {Two-stage: sun $\rightarrow$ cranks. Three-stage: offset input $\rightarrow$ compound gear $\rightarrow$ cranks.};
\end{scope}

\node[fine,anchor=west,text=ink]at(.05,-.20)
 {$W$: wheel \quad $E$: eccentric bearing \quad $S$: shaft-support bearing \quad $M$: main bearing};
\draw[motion](10.30,-.20)--(10.86,-.20);
\node[fine,anchor=west]at(10.92,-.20){drive / output};
\draw[exchange](13.12,-.20)--(13.68,-.20);
\node[fine,anchor=west]at(13.74,-.20){reaction exchange};
\end{tikzpicture}
}
\caption{Physical load paths in supported reducer configurations.
(a) A central eccentric shaft drives two wheels, whose hole contacts load
a common output sleeve and pin. Shaft supports connect to the output carrier;
main bearings connect the carrier to the stationary housing.
(b) Preliminary gearing drives multiple floating cranks sharing the wheels
and carrier. One crank is sectioned; its eccentric and support bearings are
distinct interfaces. Bearing symbols indicate mechanical connections.
Axial spacing and transverse multiplicity are schematic.
Arrows identify drive/output and reaction exchange; their directions
do not prescribe the computed gear motion or contact force.}
\label{fig:load_paths}
\end{figure}

\subsection{Geometry and deformation within the common assembly}

At each interface, current component positions and the declared
surface geometry determine the potential contacts. Manufacturing
deviations, assembly clearance, and structural displacement change
their separation. A tilt also changes separation across the face width.
These changes determine which contacts carry load and where their
resultants act. The resulting forces and moments then alter component
positions through the common assembly. Contact identity and position
are consequently required throughout the calculation, even when the
quantity of interest is a single output angle or torque.

Structural deformation transmits an influence between contacts as
well as between components. Loading one tooth can displace another
through the wheel body; two eccentric journals load the same shaft;
two wheel holes act on the same output pin. A multi-site compliance
description preserves these interactions through its cross terms.
In the shared-pin case, the influence on the second contact depends
on the pin supports, the axial separation of the wheel stations, and
the current contact normals. Separate scalar contact springs alone
do not retain this structural transfer. Resolving a floating sleeve
and its inner contact provides a further modelling choice when the
inner fit must be represented explicitly.

The framework accommodates local elastic laws, condensed static
compliances, and retained elastic coordinates. These representations
have distinct roles. A local law describes the deformation allocated
to that contact; a static compliance transfers displacements between
sites under a specified support condition; retained coordinates also
carry structural inertia and damping. A physical port consists of
selected displacement components at an interface and their conjugate
force components. Displacement traces and their
work-conjugate force maps connect the structural representation to
the contact interfaces. The same maps must enter the assembled tangent.
Geometry shared between contact, inertia, and structural assets helps
maintain consistent locations and dimensions, while explicit elastic
allocation prevents the same deformation from being counted in both
a local law and a body model. Selected shaft and seat paths instead
apply displacement offsets after a mechanical update; their update
interval forms part of the approximation. Section~\ref{sec:formulation}
gives the mathematical definitions and allocation rules.

\begin{figure}[pos=tbp]
\centering
\resizebox{\linewidth}{!}{\begin{tikzpicture}[x=1cm,y=1cm,
 box/.style={draw=ink!65,rounded corners=2pt,fill=pale,align=center,
   minimum height=1.15cm,text width=3.4cm,inner sep=6pt,outer sep=0pt,
   font=\small\relax,
   execute at begin node={\hyphenpenalty=10000\exhyphenpenalty=10000\relax}},
 arrow/.style={-{Stealth[length=2mm]},draw=ink,line width=.7pt},
 note/.style={font=\scriptsize\relax,text=muted,align=center,fill=white,inner sep=2pt}]

\node[box] (spec) at (2,5.4)
 {Assembly definition\\\scriptsize\relax Topology, geometry, coordinates};
\node[box] (local) at (7,5.4)
 {Local contact\\\scriptsize\relax Geometry, force, internal history};
\node[box] (global) at (12,5.4)
 {Global residual\\\scriptsize\relax Inertia and mapped forces};
\node[box] (flex) at (2,2.6)
 {Structural representation\\\scriptsize\relax Coordinates and port maps};
\node[box] (solve) at (7,2.6)
 {Implicit mechanical step\\\scriptsize\relax Trial solve and acceptance};
\node[box] (obs) at (12,2.6)
 {Accepted-state output\\\scriptsize\relax Motion, local loads, elapsed time};
\node[box,text width=5.2cm,fill=white] (slow) at (3.1,-.3)
 {Selected scheduled update\\\scriptsize\relax Structural, thermal or wear offsets; friction};
\node[box,text width=5.2cm,fill=white] (analysis) at (10.9,-.3)
 {Protocol-specific observations\\\scriptsize\relax Loading response, TE, motion, local stress};

\draw[arrow] (spec.east)--(local.west);
\draw[arrow] (local.east)--(global.west);
\draw[arrow] (spec.south)--(flex.north);
\draw[arrow] (flex.east)--(solve.west);
\draw[arrow] (global.south)--(12,4.05)--(7,4.05)--(solve.north);
\draw[arrow] (solve.east)--(obs.west);
\node[note,anchor=south] at (9.5,2.66) {accepted};
\draw[arrow,accent] ([xshift=-10mm]solve.north)--
                   ([xshift=-10mm]local.south);
\node[note,text=accent,anchor=east] at (5.8,3.75)
 {trial motion};

\coordinate (observed) at (12,1.35);
\draw[draw=ink,line width=.7pt] (obs.south)--(observed);
\fill[ink] (observed) circle (1.1pt);
\draw[arrow] (observed)--(12,.65)-|([xshift=11mm]analysis.north);
\draw[arrow] (observed)--(3.1,1.35)--(slow.north);
\draw[arrow,dashed,teal] (slow.west)--(-.9,-.3)--(-.9,6.95)
                         --(7,6.95)--(local.north);
\node[note,text=teal,anchor=south] at (3.2,7.02)
 {geometry / constitutive inputs};
\end{tikzpicture}
}
\caption{Common physical and computational organization. Topology closes
the motion and reaction paths; geometry and structural deformation
determine contact forces within the assembly. Accepted-state observations
retain local identities and provide inputs to separately prescribed
performance protocols. Scheduled thermal or wear updates return through
their declared geometric or constitutive channels; these outer updates
are distinct from the implicit mechanical iteration.}
\label{fig:framework}
\end{figure}

\subsection{A common mechanical state and distinct operating protocols}

The common state links component motion to local contact information
and history. Within an implicit mechanical step, repeated force and
tangent evaluations are trial calculations of that step. They do not
represent repeated physical loading events. After step acceptance,
observations use the accepted configuration and actual elapsed time.
This distinction is required when local histories influence subsequent
forces or when observations accumulate load and sliding measures.
Section~\ref{sec:solution} describes how trial evaluations, acceptance,
and observation are organized.

Performance quantities are obtained by prescribing a suitable
experiment on this assembly. A torque-loading protocol with a specified
input constraint provides angular compliance and branch response.
A rotational protocol provides transmission error relative to the
nominal kinematic ratio over a declared angular interval. Dynamic
excitation and a defined measurement location provide component or
housing response. Local pressure recovery uses the force and geometry
of a selected contact from the corresponding mechanical state.
The model can therefore connect several characteristics through a
common physical explanation, while each measurement retains its own
boundary conditions, loading history, and sampling interval.

For the implemented slower processes, accepted observations supply
load integrals, sliding-weighted loads, higher load moments, and
sampled peaks as separate quantities. A prescribed thermal, wear, or
fatigue model then selects the statistic it requires. When a supported
process changes the subsequent mechanics, it returns through a declared
clearance, surface-offset, or friction channel. Manufacturing offsets
and service increments remain separately represented. Channel
compatibility, spatial resolution, and update timing determine which
processes can be combined; a local stress or fatigue calculation does
not itself introduce a new mechanical feedback. The present numerical
studies use the mechanical and pressure-recovery paths, while these
outer processes explain how the retained contact information can be
consumed without changing its physical identity.

\subsection{Modelling choices and the numerical study}

Local discretization and global coordinate selection control different
parts of the approximation. Increasing the number of face-width
slices resolves a load distribution without creating an independent
rigid body for each slice. Reducing a structure at physical ports
changes its internal representation while retaining the interface
locations needed by the contact model. Their errors must therefore
be assessed separately. GPU batching~\cite{m_warp2022} organizes repeated
local evaluations with their contact identities intact; it does not
determine which physical coordinates or compliance terms should be
retained. Table~\ref{tab:scope} summarizes these modelling decisions
in terms of the interaction they preserve and the inputs needed to
interpret the resulting observations.

\begin{table}[pos=tbp]
\caption{Modelling choices within the common assembly. The rows describe
alternative representations and compatible extensions, rather than an
unrestricted set of simultaneous options.}
\label{tab:scope}
\centering\small
\setlength{\tabcolsep}{4pt}
\begin{tabularx}{\linewidth}{@{}>{\raggedright\arraybackslash}p{.17\linewidth}YYY@{}}
\toprule
Modelling choice & Retained interaction & Required inputs &
Observation and interpretation\\
\midrule
Supported component motion &
Contact forces change centre position and alignment through bearing reactions &
Topology, free coordinates, bearing locations, clearance/preload, and installation &
Reactions and alignment changes in the retained directions\\
\addlinespace[2pt]
Distributed local contact &
Contact recruitment, position of the resultant, and face-width load variation &
Final surface geometry, local constitutive law, and contact discretization &
Individual loads and transmitted moments; geometric and integration resolution matter\\
\addlinespace[2pt]
Static cross-compliance &
A load at one site changes separation at other sites &
Influence coefficients, loading/displacement sites, and support conditions &
Quasistatic load redistribution; no additional structural dynamics\\
\addlinespace[2pt]
Retained elastic coordinates &
Dynamic structural transfer between physical interfaces &
Mass, stiffness, damping, retained basis, and displacement/force traces &
Port motion and frequency response within the retained bandwidth\\
\addlinespace[2pt]
Accepted-state observation &
Motion, local load, history, and elapsed time remain associated &
Operating protocol, contact identity, sampling window, and cycle convention &
Branch response, TE, load statistics, and state-specific pressure recovery\\
\addlinespace[2pt]
Scheduled geometric or constitutive update &
A supported slow process changes the subsequent contact problem &
Physical update law, calibrated parameters, spatial map, and update interval &
Evolution under the declared process and compatible feedback channels\\
\bottomrule
\end{tabularx}
\end{table}

The numerical studies address three linked questions about this
organization. Section~\ref{sec:ports_example} asks which structural
responses survive a change of coordinates and which change when the
basis is truncated, using a fixed parent model and physical ports.
Section~\ref{sec:coupling_example} asks whether a shared structural
path can transfer a load change between two contacts when the second
wheel is held at a fixed pose. Its controlled cross-compliance
comparison isolates that mechanism. Section~\ref{sec:applications}
then uses the double-wheel cycloidal example to relate an assembly
loading response to aggregate contact statistics and normalized local
pressure under explicitly identified protocols. Together these
studies examine structural representation, cross-contact influence,
and the preservation of local information through an assembly
calculation. Their reference problems and operating conditions are
specified individually in the following sections; the detailed geometry
of the assembly application is outside the public disclosure.

\section{Coupled mechanical model of the transmission}
\label{sec:formulation}

The mechanical formulation follows the path identified in
Section~\ref{sec:framework}: body motion places the surfaces, their
separation determines local reactions, and the paired reaction maps
assemble those reactions into the equations governing the same motion.
The independent coordinates $\boldsymbol q$ contain the selected rigid
and elastic degrees of freedom. Contact locations and pressure or slice
unknowns are evaluated within the associated local problems; they are
not automatically additional integrated body coordinates.

\subsection{Coordinates and motion of the load-carrying bodies}

Connectivity determines which bodies share an interface; the coordinate
definition determines which relative motions that interface can constrain.
A bearing-supported crank has independent motion that is resolved by its
bearing reactions, whereas a prescribed revolute connection imposes its
kinematic relation directly. Assembly phase is established from the selected
gear geometry before either representation is loaded. In particular,
opposed eccentric journals and conjugate wheel-tooth phases are separate
parts of that initial configuration.

Body coordinates are selected from translations, rotations, and retained
elastic coordinates. Locking a coordinate removes it from the system;
allowing a crank to translate and tilt leaves its position and orientation
to be determined by bearing contact. For a reduced revolute connection,
the child origin and absolute angle satisfy
\begin{equation}
 \boldsymbol o_b=\boldsymbol o_p+\boldsymbol R_z(\theta_p)\boldsymbol d_{pb},
 \qquad \theta_b=\theta_p+\vartheta_b,
 \label{eq:parent_kinematics}
\end{equation}
where $\vartheta_b$ is a relative angle. A floating crank supported by elastic
bearings instead retains independent coordinates. These alternatives represent
different mechanical constraints.

For the planar rotation and small-tilt body model, a material surface point has
the applicable rigid and elastic contributions
\begin{equation}
 \boldsymbol x_b(\boldsymbol q,\boldsymbol\xi)
 =\boldsymbol o_b+\boldsymbol R_z(\theta_b)\boldsymbol r_b(\boldsymbol\xi)
 +\boldsymbol\theta_{\perp,b}\mathbin{\times}
             \boldsymbol R_z(\theta_b)\boldsymbol r_b(\boldsymbol\xi)
 +\boldsymbol H_b(\boldsymbol\xi)\boldsymbol\eta_b.
 \label{eq:surface_motion}
\end{equation}
Here $\boldsymbol\xi$ is a local material coordinate; the selected body and
endpoint model determine which terms are present. For a stationary reduced
housing, $\boldsymbol H_b\boldsymbol\eta_b$ supplies interface deformation,
without implying a floating flexible-body formulation. Actual axial offsets
of meshes and bearing planes enter the point lever arms. They therefore affect
both tilt-induced gaps and the moments generated by contact forces.

The distinct reaction paths and axial stations are summarized in
\figref{fig:load_paths}; the next subsections resolve their local interfaces.

\subsection{Tooth geometry, deviations, and contact location}

\begin{figure}[pos=tbp]
\centering
\resizebox{\linewidth}{!}{\begin{tikzpicture}[x=1cm,y=1cm,
  label/.style={font=\sffamily\fontsize{8}{9.5}\selectfont,align=center},
  box/.style={draw=ink!65,rounded corners=2pt,fill=pale,
    text width=3.0cm,minimum height=.8cm,inner sep=4pt,align=center,
    font=\sffamily\fontsize{8}{9.5}\selectfont},
  arrow/.style={-{Stealth[length=1.7mm]},draw=ink,line width=.65pt}]
\node[label] at (2.1,3.6) {(a) Local surface geometry};
\draw[teal,line width=1pt] (.6,3.0) arc[start angle=180,end angle=360,x radius=1.5,y radius=1.3];
\draw[ink,line width=1pt] (.3,.9) .. controls (1.1,1.433333) and (3.1,1.433333) .. (3.9,.9);
\node[label,text=teal] at (2.1,2.8) {Body $B$};
\node[label,text=ink] at (2.1,.45) {Body $A$};
\fill[accent] (2.1,1.7) circle (1.8pt);
\fill[accent] (2.1,1.3) circle (1.8pt);
\draw[<->,accent,line width=.7pt] (2.1,1.3)--(2.1,1.7);
\node[label,text=accent,anchor=west] at (2.2,1.5) {$g$};
\draw[arrow] (2.1,1.3)--(2.1,.75);
\node[label,anchor=east] at (2.05,.85) {$\boldsymbol n$};
\draw[arrow] (2.1,1.3)--(2.7,1.3);
\node[label,anchor=south] at (2.65,1.35) {$\boldsymbol t$};

\node[label] at (7.0,3.6) {(b) Geometry and deformation};
\node[box] (pose) at (7,2.7) {Surface motion\\$\boldsymbol x_A,\ \boldsymbol x_B$};
\node[box] (gap) at (7,1.5) {Effective gap\\$g=g^{\rm geom}+u^{\rm off}$};
\node[box] (force) at (7,.25) {Contact reaction\\$F_n=\mathcal F([{-g}]_+)$};
\draw[arrow] (pose.south)--(gap.north);
\draw[arrow] (gap.south)--(force.north);

\node[label] at (12,3.6) {(c) Force--motion pairing};
\node[box] (coordinates) at (12,2.7) {Selected coordinates\\$\boldsymbol q$};
\node[box] (increment) at (12,1.5) {Material-point increment\\$\delta\boldsymbol u=\boldsymbol J\delta\boldsymbol q$};
\node[box] (reaction) at (12,.25) {Generalized reaction\\$\boldsymbol Q=\boldsymbol J^{\mathsf T}\boldsymbol f$};
\draw[arrow] (coordinates.south)--(increment.north);
\draw[arrow] (increment.south)--(reaction.north);
\end{tikzpicture}
}
\caption{Schematic local contact description and motion--force pairing.
(a) Two representative surface segments define a signed separation and
local directions. (b) Relative geometry and the allocated deformation
determine the contact reaction. (c) Material-point increments and their
work-conjugate force map connect this reaction to the assembly.
The surfaces and spacing are illustrative and are not component drawings.}
\label{fig:geometry}
\end{figure}

Local geometry is retained explicitly. A canonical involute flank is
\begin{equation}
 \boldsymbol c_{\mathrm{inv}}(u)
 =r_b\begin{bmatrix}\cos u+u\sin u\\ \sin u-u\cos u\end{bmatrix},
 \qquad
 \boldsymbol c_{\mathrm{mod}}(u)
 =\boldsymbol c_{\mathrm{inv}}(u)+e(u)\boldsymbol n(u).
 \label{eq:involute_geometry}
\end{equation}
Tooth index, flank orientation, thickness, and body rotation place this curve
in the assembly. Modification changes its position and, where evaluated
analytically, its derivatives. The involute contact path finds intersections
along candidate flank normals within admissible profile intervals, retaining
the contact location and force lever arm.

The implemented nominal cycloidal profile can be written as
\begin{equation}
 \begin{aligned}
 \boldsymbol c_{\mathrm{cyc}}(\phi)&=
 R_z\begin{bmatrix}
 \sin\phi-(K_1/z_b)\sin(z_b\phi)\\
 \cos\phi-(K_1/z_b)\cos(z_b\phi)
 \end{bmatrix}
 +\frac{r_c(\phi)}{D(\phi)}
 \begin{bmatrix}K_1\sin(z_b\phi)-\sin\phi\\
 K_1\cos(z_b\phi)-\cos\phi\end{bmatrix},
 \\
 D(\phi)&=\sqrt{1+K_1^2-2K_1\cos(z_g\phi)}.
 \end{aligned}
 \label{eq:cycloid_geometry}
\end{equation}
Here $R_z$ is the pin-circle radius, $K_1$ the eccentricity coefficient,
$z_b$ and $z_g$ the pin and wheel tooth counts, and $r_c$ the profile-generating
radius including the prescribed modification. The nominal generating
eccentricity is $e_g=R_zK_1/z_b$; an explicitly specified assembly
eccentricity remains a separate input. The discrete path resamples the final profile approximately
uniformly in arc length and searches the available pins against it.
A local tangent projection refines the selected station. An alternative
analytic path solves $(\boldsymbol c-\boldsymbol p)\cdot\boldsymbol c'=0$
within a bracket initialized from several candidates. Search geometry and
constitutive curvature are specified together for the chosen contact path.
The final modified polygon determines the discrete location; a stiffness
model using a separate curvature table retains that table's approximation.

The effective normal gap is
\begin{equation}
 g_c=g_c^{\mathrm{geom}}+e_c^{\mathrm{man}}+e_c^{\mathrm{assembly}}
       +u_c^{\mathrm{off}},\qquad \delta_c=\max(-g_c,0).
 \label{eq:gap}
\end{equation}
Positive gap denotes separation. The deviation terms include only quantities
not already embedded in surface coordinates: individual pin-center errors,
wheel normal deviations, clearance, or structural displacement offsets.
The discrete cycloidal search also retains an activation clearance: projected
distance is used inside the permitted local interval, while an out-of-interval
projection is tested conservatively using the sampled point. The activation
clearance and indentation distance can therefore differ at a clamped projection.

\subsection{Normal compliance, friction, and integration across the width}

Contact combines a family-specific elastic law and the declared damping.
For a gear slice with positive indentation,
\begin{equation}
 F_{n,c}=\max\!\left(0,\mathcal F_c(\delta_c)
       +c_c r_c^{\mathrm d}(\delta_c)\dot\delta_c\right),
 \qquad r_c^{\mathrm d}(\delta)=\min(\delta/\delta_{r,c},1),
 \label{eq:normal_force}
\end{equation}
and an inactive slice carries no force. The ramp regularizes damping at
engagement. Point contact, roller laminae, and load-linearized gear contact
use different functions $\mathcal F_c$; a coefficient multiplying $\delta^p$
has units $\mathrm{N\,m}^{-p}$.

The current involute real-stiffness model closes the local elastic series
relation directly:
\begin{equation}
 \delta_c=\frac{F_c}{k_{s,A}}+\frac{F_c}{k_{s,B}}+\delta_H(F_c),
 \qquad
 k_{t,c}=\left(\frac1{k_{s,A}}+\frac1{k_{s,B}}
                    +\frac{d\delta_H}{dF_c}\right)^{-1}.
 \label{eq:series_contact}
\end{equation}
The first two terms are the selected tooth and local-body contributions;
$\delta_H$ is the implemented line-contact indentation. A safeguarded scalar
iteration resolves its load dependence. The cycloidal implementation also
supports a two-body line-contact secant based on its stored reference load;
calibrated power laws remain available for other interfaces. These choices
determine the load dependence of the instantaneous contact tangent.

For an in-plane normal $(n_x,n_y)$, relative small tilt and prescribed
misalignment give
\begin{equation}
 e_{\mathrm{skew}}(z)
 =z(\Delta\theta_y n_x-\Delta\theta_x n_y)+z\tan\psi.
 \label{eq:skew}
\end{equation}
The last term is the implemented normal-direction skew approximation.
Crowning and axial offsets supply additional terms where declared.
Each width slice is activated separately, so an edge can carry load while
the midplane is open. With slice forces already weighted by their widths,
\begin{equation}
 \boldsymbol F=\sum_s\boldsymbol f_s,\qquad
 \boldsymbol M_O=\sum_s(\boldsymbol x_s-\boldsymbol O)\mathbin{\times}\boldsymbol f_s.
 \label{eq:width_resultant}
\end{equation}
This retains the bending moment generated by unequal end loads, which is
lost when a complete face is replaced by one force at its midplane.

For velocity-dependent friction,
\begin{equation}
 F_{t,c}=-F_{n,c}\mu_{\mathrm{reg}}(v_{t,c}),\qquad
 \mu_{\mathrm{reg}}(v)v\geq0.
 \label{eq:friction}
\end{equation}
The velocity is the relative velocity of the two material surfaces at the
contact. Differentiating the searched closest-point position would also
include migration along a surface, which is not material sliding.
The alternative anchor model stores tangential elastic displacement in
body-fixed coordinates, limits its force by the friction bound, and projects
the anchors during slip. Its history enters the accepted-step treatment
in Section~\ref{sec:solution}.

\subsection{Distributed bearing support and spatial load sharing}

Bearings determine the positions of the floating wheels, cranks, and carrier.
Their reactions are calculated from individual rolling elements rather than
a prescribed radial stiffness. For ball $j$, let $a_{r,j}$ and $a_{z,j}$ be
the relative radial and axial components between groove-curvature centers.
Translation, ring tilt, clearance, and preload determine
\begin{equation}
 \delta_j=\max\!\left(\sqrt{a_{r,j}^2+a_{z,j}^2}-A-c_b,0\right),
 \quad \alpha_j=\operatorname{atan2}(a_{z,j},a_{r,j}),
 \quad F_j=K_j(\alpha_j)\delta_j^{3/2}.
 \label{eq:ball_bearing}
\end{equation}
Here $A$ is the reference groove-center separation. The working angle changes
both load direction and contact stiffness. Inner and outer point-contact
stiffnesses combine as
$K_j=(K_{i,j}^{-2/3}+K_{o,j}^{-2/3})^{-3/2}$
when they transmit the same load. Generalized reactions follow the derivative
of the same normal deformation used in Eq.~\eqref{eq:ball_bearing}.

Cylindrical and crossed rollers are sliced along their generators. A
cylindrical roller has a radial normal; a crossed-roller row alternates the
axial component through opposite contact angles. Slice approach includes
normal ring displacement, tilt lever arms, crowning, and clearance.
Independent lamina laws provide an economical representation. The supported
influence-coefficient alternative couples slice forces through a finite-width
elastic matrix. Its fixed-width subproblem is
\begin{equation}
 \boldsymbol w=\boldsymbol C_\ell\boldsymbol f-\boldsymbol b,\qquad
 \boldsymbol f\geq0,\quad\boldsymbol w\geq0,\quad
 \boldsymbol f^{\mathsf T}\boldsymbol w=0,
 \label{eq:bearing_ic}
\end{equation}
where $\boldsymbol b$ is the nonnegative geometric approach supplied by this
path. The matrix is a displacement-per-force compliance; contributions from
both races add when included. Contact half-width may be fixed or updated from
the total line load. Coupled compliance permits nonuniform finite-width
loading even under uniform approach. A precomputed edge weight is a separate
approximation to that response.

Tapered rollers additionally resolve cone geometry and, where enabled, rib
contact. Forces and moments are summed at the actual bearing planes. Paired
supports resist axial force and overturning moment through their separation
as well as their contact angles. Preload changes the loaded element set
before external loading; subsequent support deformation redistributes both
bearing and gear loads.

\subsection{Structural operators, contact ports, and coupled deformation}
\label{sec:structural_ports}

A component transmits load between spatially separated interfaces. A tooth
force can deform a wheel hole, and a housing reaction at one pin can change
the support of another. These interactions require a structural operator
connecting the interfaces, together with a specified mounting condition.
Recent substructuring studies combine interface reduction with detailed
joint and support models~\cite{m_zhang2024_substruct,m_hippold2025_interfaces}.
The present implementation provides static influence coefficients and retained
dynamic coordinates as distinct representations. Figure~\ref{fig:structural_ports}
follows the latter from a mounted component to the contact calculation;
the same work-conjugacy principle also defines static projection.

\begin{figure}[pos=tbp]
\centering
\resizebox{\linewidth}{!}{\begin{tikzpicture}[x=1cm,y=1cm,
  font=\sffamily\fontsize{8}{9.5}\selectfont,
  line cap=round,line join=round,
  metal/.style={draw=ink,line width=.7pt,fill=pale},
  small/.style={font=\sffamily\fontsize{7}{8.5}\selectfont,text=muted},
  title/.style={font=\sffamily\bfseries\fontsize{9}{10.5}\selectfont,text=ink},
  disp/.style={draw=teal,line width=.8pt,-{Stealth[length=1.6mm]}},
  load/.style={draw=accent,line width=.8pt,-{Stealth[length=1.6mm]}},
  lead/.style={draw=muted,line width=.5pt},
  flow/.style={draw=ink,line width=.7pt,-{Stealth[length=1.6mm]}}]
\path[use as bounding box](0,0)rectangle(16,7.3);
\draw[muted!35](5.32,.15)--(5.32,6.84) (10.68,.15)--(10.68,6.84);
\node[title,anchor=west]at(.05,7.03){(a) Component and mounting};
\node[title,anchor=west]at(5.50,7.03){(b) Structural reduction};
\node[title,anchor=west]at(10.86,7.03){(c) Contact and system};

\begin{scope}[shift={(2.5,4.9)}]
 \fill[pale,even odd rule](0,0)circle(1.52)(0,0)circle(1.20);
 \draw[ink,line width=.8pt](0,0)circle(1.52)(0,0)circle(1.20);
 \foreach \k in {0,...,51}{
   \fill[teal]({\k*(360/52)}:1.20)circle(.027);}
 \foreach \k in {0,...,17}{
   \begin{scope}[rotate={360*\k/18}]
    \draw[ink,line width=.4pt,fill=white](1.382,-.027)rectangle(1.436,.027);
   \end{scope}}
 \fill[accent](1.20,0)circle(.058);
 \fill[teal]({8*(360/52)}:1.20)circle(.058);
 \draw[load](.30,0)--(1.08,0);
 \node[anchor=south,text=accent]at(.63,.12){$f_0$};
 \node[small,anchor=north west]at(1.25,-.05){$p_0$};
 \draw[disp]({8*(360/52)}:1.23)--({8*(360/52)}:1.67);
 \node[anchor=west,text=teal]at(1.04,1.39){$u_8$};
 \node[small,anchor=south east]at(.57,.86){$p_8$};
 \node[small,align=center]at(0,-.53){52 pin ports\\$\{x,y,z\}$ at each port};
\end{scope}
\node[small,anchor=west]at(.15,3.06){Uniform annulus; supports at the lower face};
\draw[metal](.63,2.10)rectangle(1.21,2.70);
\draw[metal](3.79,2.10)rectangle(4.37,2.70);
\draw[lead,densely dashed](1.21,2.39)--(3.79,2.39);
\node[small,fill=white,inner sep=1.2pt]at(2.5,2.39){bore};
\foreach \x in {.92,4.08}{
 \draw[ink,line width=.6pt](\x,2.10)--++(0,-.10)--++(-.07,-.07)
   --++(.14,-.10)--++(-.14,-.10)--++(.14,-.10)--++(-.07,-.07)--++(0,-.10);
 \draw[ink,line width=.6pt](\x-.22,1.46)--(\x+.22,1.46);
 \foreach \d in {-.20,-.10,0,.10,.20}{\draw[lead](\x+\d,1.46)--++(-.07,-.09);}}
\node[small,align=center]at(2.5,1.81){18 mounting springs\\$k_{\rm bolt}$ included in $\boldsymbol K$};
\node[align=center]at(2.5,.67){$\boldsymbol K_f^{-1}$: supported solve\\
 $\boldsymbol C_p=\boldsymbol P_f\boldsymbol K_f^{-1}\boldsymbol P_f^{\mathsf T}$};

\node[align=center]at(8,6.33){$\boldsymbol q_{\rm CB}=[\boldsymbol u_b;\boldsymbol\zeta]$\\
 $156$ boundary components\\$+\,20$ fixed-interface modes};
\draw[flow](8,5.84)--(8,5.48);
\node[align=center]at(8,5.14){$\boldsymbol K_r\boldsymbol\Phi=
 \boldsymbol M_r\boldsymbol\Phi\boldsymbol\Omega^2$};
\draw[ink,line width=.6pt](5.87,4.62)--(5.72,4.62)--(5.72,3.10)--(5.87,3.10);
\draw[ink,line width=.6pt](7.25,4.62)--(7.40,4.62)--(7.40,3.10)--(7.25,3.10);
\foreach \x in {5.98,6.27,6.56,6.85,7.14}{
 \draw[teal!65,line width=3.3pt](\x,3.30)--(\x,4.43);}
\node[small]at(6.56,2.88){$\boldsymbol\Phi:\;r\times n_\eta$};
\draw[flow](7.67,3.86)--(8.28,3.86);
\node[small,anchor=south]at(7.98,4.00){$\boldsymbol S_b$};
\draw[ink,line width=.6pt](8.69,4.30)--(8.54,4.30)--(8.54,3.42)--(8.69,3.42);
\draw[ink,line width=.6pt](10.07,4.30)--(10.22,4.30)--(10.22,3.42)--(10.07,3.42);
\foreach \x in {8.80,9.09,9.38,9.67,9.96}{
 \draw[teal,line width=3.3pt](\x,3.58)--(\x,4.14);}
\node[small]at(9.38,2.88){$\boldsymbol H:\;156\times n_\eta$};
\node[align=center]at(8,2.14){One row: one physical port component\\
 One column: one retained coordinate};
\node[small,align=center]at(8,1.32){Contact resolution is preserved;\\
 the structural subspace is selected separately};
\node[align=center]at(8,.52){$\boldsymbol M_\eta=\boldsymbol I,\quad
 \boldsymbol K_\eta=\boldsymbol\Omega^2$};

\node[small,align=center]at(13.4,6.26){Port displacement changes the gap;\\
 the contact load returns to the same structure};
\draw[teal,line width=.85pt,fill=teal!9]
 (11.84,3.37)--(12.31,3.56)..controls(12.90,3.73)and(13.11,4.22)..(13.27,4.65)
 --(12.86,4.94)..controls(12.71,4.43)and(12.42,4.12)..(11.84,4.02)--cycle;
\draw[metal](13.99,4.39)circle(.69);
\draw[ink,line width=.6pt](14.68,4.15)--(15.15,4.15)
 --(15.15,4.62)--(14.68,4.62);
\fill[accent](13.33,4.20)circle(.055);
\draw[load](13.33,4.20)--(13.98,4.43);
\node[anchor=south,text=accent]at(13.82,4.70){$f_g$};
\node[small]at(12.25,3.15){wheel flank};
\node[small]at(14.13,3.49){pin / housing port};
\draw[disp](11.31,1.71)--(11.31,5.48)--(14.02,5.48)--(14.02,5.18);
\node[anchor=south,text=teal]at(12.68,5.55){$\boldsymbol u_g=\boldsymbol H_g\boldsymbol\eta$};
\draw[load](15.20,4.39)--(15.80,4.39)--(15.80,1.73)--(14.89,1.73);
\node[anchor=west,text=accent,rotate=90]at(15.57,2.19){$\boldsymbol H_g^{\mathsf T}\boldsymbol f_g$};
\node[align=center]at(13.33,1.85){Integrated coordinates\\
 $\boldsymbol q_s=[\boldsymbol q_r;\boldsymbol\eta]$};
\draw[disp](12.00,1.73)--(11.31,1.73);
\node[align=center]at(13.38,.71){$\boldsymbol q_e=\boldsymbol T_e\boldsymbol q_s$\\
 $\boldsymbol Q_s=\boldsymbol T_e^{\mathsf T}\boldsymbol Q_e,\quad
 \boldsymbol J_s=\boldsymbol T_e^{\mathsf T}\boldsymbol J_e\boldsymbol T_e$};
\end{tikzpicture}
}
\caption{From a mounted structure to coupled contact evaluation.
(a) An idealized annular housing has 52 pin locations and 18 discrete
mounting supports; each three-dimensional pin port contains three
translation components. The lower section illustrates the nominal lower-face
mounting region; the constructor assigns nearby mesh nodes to the springs.
(b) Craig--Bampton reduction retains physical boundary
coordinates; a second basis maps all 156 boundary components to a selectable
number of generalized coordinates. (c) The same mapping supplies contact
displacement, returns force, and contracts the complete tangent.
The geometry and deformation drawings are explanatory schematics, not
finite-element meshes or calculated mode shapes.}
\label{fig:structural_ports}
\end{figure}

\subsubsection{Static operators and their physical boundaries}

For nodal displacement $\boldsymbol u$, a normal contact port samples
$u_{p,c}=\sum_a N_a(\boldsymbol x_c)\boldsymbol n_c^{\mathsf T}\boldsymbol u_a$.
In the background-grid component path, trilinear interpolation and the
normalized port direction define one row of $\boldsymbol P$; its transpose
distributes the corresponding force to nodes. The nodal housing ports used
below instead select Cartesian components at their assigned mesh vertices.
Restricting this projection to the supported free coordinates gives
\begin{equation}
 \boldsymbol C_b=\boldsymbol P_f\boldsymbol K_f^{-1}\boldsymbol P_f^{\mathsf T},
 \qquad \boldsymbol u_p=\boldsymbol C_b\boldsymbol f_p.
 \label{eq:body_compliance}
\end{equation}
Here the inverse denotes back-substitution using a stored factorization.
The discretization, supports, port locations, and directions belong to the
operator definition. A periodic table can replace repeated solves when the
ports follow a prescribed contact band. Self-compliance is interpolated
separately from cross-compliance to avoid smoothing its narrow diagonal peak.

A freely supported wheel instead requires removal of its six rigid-body
motions. The implemented inertia-relief solve augments the stiffness by
mass-weighted rigid modes $\boldsymbol R_{\rm rb}$ and imposes
$\boldsymbol R_{\rm rb}^{\mathsf T}\boldsymbol M\boldsymbol u=\boldsymbol0$.
The multiplier balances the rigid component of the applied load; the returned
field describes deformation relative to this frame. It is not equivalent
to clamping the wheel bore. Grouping tooth and support-hole ports exposes
the available coupling:
\begin{equation}
 \begin{bmatrix}\boldsymbol u_t\\\boldsymbol u_h\end{bmatrix}
 =\begin{bmatrix}\boldsymbol C_{tt}&\boldsymbol C_{th}\\
                  \boldsymbol C_{ht}&\boldsymbol C_{hh}\end{bmatrix}
   \begin{bmatrix}\boldsymbol f_t\\\boldsymbol f_h\end{bmatrix}.
 \label{eq:multiport_body}
\end{equation}
If the declared support ports are held fixed, eliminating their reactions
on the nonsingular constrained subspace gives
\begin{equation}
 \boldsymbol C_{t\mid h}=\boldsymbol C_{tt}
 -\boldsymbol C_{th}\boldsymbol C_{hh}^{-1}\boldsymbol C_{ht}.
 \label{eq:port_condensation}
\end{equation}
This is a conditional body compliance; bearing indentation remains a
separate physical contribution. The complete stored matrix does not imply
that every block is simultaneously active in the time-domain model.

The supported quasistatic feedback uses selected tooth cross-compliances,
condensed crank bending, hole-wall response, and carrier contributions to
update contact or race offsets after accepted steps:
\begin{equation}
 \boldsymbol u_{k+1}^{\rm off}
 =(1-\omega)\boldsymbol u_k^{\rm off}
   +\omega\mathcal C(\boldsymbol f_k),\qquad 0<\omega\leq1.
 \label{eq:weak_coupling}
\end{equation}
Relaxation and update interval control this delayed feedback. The configured
annular-plate term is instead a constant offset calculated from the specified
output torque. These offsets carry no structural inertia and remain fixed
within a mechanical trial; they cannot represent the dynamic response of
explicit elastic coordinates.

\subsubsection{Dynamic reduction with a retained interface map}

For dynamic deformation, finite-element mass and stiffness are partitioned
into boundary and interior coordinates. Craig--Bampton reduction uses
constraint modes and fixed-interface modes $\boldsymbol\Phi_i$~\cite{m_craig1968}:
\begin{equation}
 \begin{aligned}
 \begin{bmatrix}\boldsymbol u_b\\\boldsymbol u_i\end{bmatrix}
 &=\underbrace{\begin{bmatrix}\boldsymbol I&\boldsymbol0\\
 -\boldsymbol K_{ii}^{-1}\boldsymbol K_{ib}&\boldsymbol\Phi_i
 \end{bmatrix}}_{\boldsymbol T_{\rm CB}}
 \boldsymbol q_{\rm CB},\\
 \boldsymbol M_r&=\boldsymbol T_{\rm CB}^{\mathsf T}\boldsymbol M
                         \boldsymbol T_{\rm CB},\qquad
 \boldsymbol K_r=\boldsymbol T_{\rm CB}^{\mathsf T}\boldsymbol K
                         \boldsymbol T_{\rm CB}.
 \end{aligned}
 \label{eq:cb_reduction}
\end{equation}
Physical interface components remain independent coordinates at this stage.
For the mounted annulus, discrete translational mounting springs enter
$\boldsymbol K$ before reduction; replacing them by a clamped circumference
changes the model. The subsequent port reduction requires positive-definite
$\boldsymbol K_r$ and rejects an ungrounded block. It therefore describes
the supported stationary structure, rather than an unrestricted floating
flexible body. Fixed rigid axes are eliminated from the system, whereas a
stationary flexible housing retains its elastic coordinates; grounding does
not make that component rigid.

A second generalized eigenproblem,
$\boldsymbol K_r\boldsymbol\Phi_T=
\boldsymbol M_r\boldsymbol\Phi_T\boldsymbol\Omega_T^2$,
selects a mass-normalized subspace. With retained and optional residual
attachment modes collected in $\boldsymbol\Phi$, define
$\boldsymbol H_g=\boldsymbol S_g\boldsymbol\Phi$, where
$\boldsymbol S_g$ selects the physical components of interface $g$.
The paired maps are
\begin{equation}
 \begin{aligned}
 \boldsymbol u_g&=\boldsymbol H_g\boldsymbol\eta,
 &\boldsymbol Q_\eta&=\sum_g\boldsymbol H_g^{\mathsf T}\boldsymbol f_g,\\
 \dot{\boldsymbol\eta}^{\mathsf T}\boldsymbol Q_\eta
 &=\sum_g\dot{\boldsymbol u}_g^{\mathsf T}\boldsymbol f_g.
 \end{aligned}
 \label{eq:port_work_pair}
\end{equation}
Thus a force at one port excites coordinates shared by the other ports.
Increasing contact resolution need not increase the number of structural
coordinates by the same amount. The reduced mass is the identity; stiffness
is diagonal, while projected damping is retained as a full matrix when
it is not proportional.

Truncation is assessed at the ports. For their stacked selector
$\boldsymbol S_b$, the omitted static response is
\begin{equation}
 \boldsymbol R_p=\boldsymbol S_b\boldsymbol K_r^{-1}\boldsymbol S_b^{\mathsf T}
 -\boldsymbol H_T\boldsymbol\Omega_T^{-2}\boldsymbol H_T^{\mathsf T}.
 \label{eq:port_residual_flexibility}
\end{equation}
The implemented reduction constructs residual attachment vectors from this
response and optionally retains them as additional dynamic coordinates.
Its hybrid decomposition also exports the remaining self-blocks for static
reconstruction. Those exported blocks are not automatically included in
the current contact runtime: comparisons therefore distinguish the actual
modal response from the separately corrected static reconstruction.
Static compliance and frequency-response convergence must both be checked;
a frequency cutoff alone does not establish contact accuracy.

\subsubsection{Assembly and allocation of elasticity}

Contact routines evaluate physical port displacement and velocity in an
expanded workspace. Let $\boldsymbol q_s=[\boldsymbol q_r;\boldsymbol\eta]$
be the integrated coordinates and let $\boldsymbol B$ insert all port maps.
Then
\begin{equation}
 \boldsymbol q_e=\underbrace{\begin{bmatrix}\boldsymbol I\\\boldsymbol B\end{bmatrix}}
 _{\boldsymbol T_e}\boldsymbol q_s,\qquad
 \boldsymbol Q_s=\boldsymbol T_e^{\mathsf T}\boldsymbol Q_e,\qquad
 \boldsymbol J_s=\boldsymbol T_e^{\mathsf T}\boldsymbol J_e\boldsymbol T_e.
 \label{eq:port_runtime_transform}
\end{equation}
The workspace adds no independent unknowns. The tangent transformation
preserves rigid--port and port--port cross-blocks, not just the structural
diagonal. It is exact for this constant coordinate map; the underlying
contact derivative must still follow the selected geometric linearization.

Finally, each elastic region is allocated once. In the hybrid tooth model,
analytical bending and shear combine with a tooth-free body operator,
and the analytical foundation term is disabled. An explicit housing or shaft
representation likewise requires removal of a duplicate static contribution.
The selected regions, boundaries, reduction order, and feedback schedule
therefore define the structural model as fully as its material constants.

\subsection{Output-pin transfer and a reproducible two-wheel coupling}

An output pin inside a wheel hole has the gap
\begin{equation}
 g_{wp}=R_h-r_p-\|\boldsymbol c_p-\boldsymbol c_h\|
                    +u_{wp}^{\mathrm{off}}.
 \label{eq:output_gap}
\end{equation}
The pin center follows the flange, whereas the hole center follows the wheel.
Their different translations, rotations, and axial planes generate approach
and force lever arms. The force on the pin points towards the hole center;
its opposite acts on the wheel. Summation about the flange axis produces
output torque without imposing equal sharing among pins.

Two wheels loading one pin provide a small reproducible example of the
framework's coupling. Its inputs are the two wheel and flange poses, the pin
and hole radii, contact lengths, wheel and support axial stations, shaft
section dimensions and material, and the contact-law coefficient. Beam
condensation supplies $C_{00},C_{11},C_{01}$ at the wheel stations; declared
inner-contact compliance contributes to the diagonal terms. The unknown
contact resultants and deflections are coupled through
\begin{equation}
 \boldsymbol C_p=
 \begin{bmatrix}C_{00}&C_{01}(\boldsymbol n_0\cdot\boldsymbol n_1)\\
 C_{01}(\boldsymbol n_0\cdot\boldsymbol n_1)&C_{11}\end{bmatrix}.
 \label{eq:pin_compliance}
\end{equation}
Aligned and opposed wheel loads therefore give different shaft-deflection
corrections even when their magnitudes are equal.

The implemented condensed calculation has three local stages. First, for
wheel $i$ with rigid-support slice approaches $d_{is}$, the output-pin power
law gives
\begin{equation}
 F_i^0=\sum_s k_s[d_{is}]_+^{10/9},\qquad
 K_i=\sum_{s:d_{is}>0}\frac{10}{9}k_s
              \max(d_{is},d_{\min})^{1/9},\qquad
 \boldsymbol K_c=\operatorname{diag}(K_0,K_1),
 \label{eq:pin_open_state}
\end{equation}
where $k_s$ is the whole-contact coefficient divided by the number of uniform
slices, $[d]_+=\max(d,0)$, and $d_{\min}=10^{-9}\,\mathrm m$ is the
implemented tangent regularization. This coefficient is calibrated as
$k=F_{\mathrm{ref}}/\delta_{\mathrm{ref}}^{10/9}$ at the specified local Hertz
reference load, preserving its units. Second, it solves the two-by-two system
\begin{equation}
 (\boldsymbol I+\boldsymbol K_c\boldsymbol C_p)\boldsymbol F^c
 =\boldsymbol F^0,\qquad
 \boldsymbol u_p=\boldsymbol C_p\boldsymbol F^c,
 \label{eq:pin_local_closure}
\end{equation}
clipping tensile corrected resultants to zero. Third, it subtracts the
resulting station deflections from slice approach, reevaluates the nonlinear
slice forces, and scatters opposite reactions to both wheels and the flange
with their axial lever arms. This is a tangent-based condensation followed
by nonlinear reevaluation, with the accuracy of that local closure assessed
separately from the global residual. The supported clearance-fit variant also
equilibrates a massless sleeve center by projection onto its radial clearance
disk.

An alternative retains explicit sleeve motion and separate inner and outer
contacts. Within its flexible shaft-to-sleeve interface, each axial slice
solves $\delta_i=\delta_i^{\mathrm{rigid}}-\sum_jC_{ij}F_j$ with its nonlinear
normal law. Inner-contact and wheel-hole forces act through the same sleeve
coordinates. This separates shaft-induced load redistribution from
clearance-driven sleeve motion, while allowing their different inertia
assumptions to be studied.

\subsection{Power-conjugate assembly and complete mechanical balance}

Let $\boldsymbol f_c$ be the Cartesian force on body $A$ in the common
spatial frame, with $-\boldsymbol f_c$ acting on body $B$.
At the selected material points, define
$\boldsymbol J_A=(\partial\boldsymbol x_A/\partial\boldsymbol q)_{\boldsymbol\xi_A}$
and similarly for $B$. Holding local material coordinates fixed gives
\begin{equation}
 \begin{aligned}
 \boldsymbol v_{\mathrm{rel}}&=(\boldsymbol J_A-\boldsymbol J_B)\boldsymbol v,\qquad
 \boldsymbol Q_c=(\boldsymbol J_A-\boldsymbol J_B)^{\mathsf T}\boldsymbol f_c,\\
 \boldsymbol v^{\mathsf T}\boldsymbol Q_c
 &=\boldsymbol v_{\mathrm{rel}}^{\mathsf T}\boldsymbol f_c.
 \end{aligned}
 \label{eq:contact_mapping}
\end{equation}
This identity transfers contact power to the system, including tilt moments
and structural ports. The point maps describe material motion at the selected
contact location; search-station migration is treated by the geometric
evaluation. Writing $\boldsymbol G_c=\boldsymbol J_A-\boldsymbol J_B$,
the smooth-branch force derivative contains two contributions:
\begin{equation}
 \frac{\partial\boldsymbol Q_c}{\partial q_j}
 =\frac{\partial\boldsymbol G_c^{\mathsf T}}{\partial q_j}\boldsymbol f_c
  +\boldsymbol G_c^{\mathsf T}
          \frac{\partial\boldsymbol f_c}{\partial q_j}.
 \label{eq:assembled_contact_tangent}
\end{equation}
The first accounts for changing moment arms and motion directions in the
mapping; the second contains the local force response, including its
selected structural coupling. The derivatives refer to the differentiable
contact description chosen for that evaluation. When a discrete search
candidate is frozen, the same product rule applies conditional on that
candidate; it does not supply the derivatives of a subsequent station
change. Keeping only a scalar indentation stiffness
would omit part of this assembly derivative. The numerical tangent follows
the approximation of the selected contact path, as specified in
Section~\ref{sec:solution}. The shared-pin moment diagnostic in
Appendix~\ref{app:moment_diagnostic} examines the force-arm contribution
in this identity.

The assembled equations are
\begin{equation}
 \boldsymbol M(\boldsymbol q)\boldsymbol a
 +\boldsymbol h(\boldsymbol q,\boldsymbol v)
 -\boldsymbol Q(\boldsymbol q,\boldsymbol v;\boldsymbol z,\boldsymbol\lambda)
 =\boldsymbol0,
 \label{eq:eom}
\end{equation}
where $\boldsymbol Q$ contains contact, elastic restoring, damping, drive,
and external-load contributions. History $\boldsymbol z$ contains the
registered contact-memory variables. Held controls $\boldsymbol\lambda$
include actuation and the geometric or constitutive fields supplied to the
current mechanical step; scheduled slow processes update their assigned
fields between mechanical windows. The mass contains the applicable
constant body blocks, parent--child inertial coupling, and reduced structural
blocks. In the supported configuration-dependent planar path, a body contributes
$\boldsymbol J_v^{\mathsf T}m\boldsymbol J_v+
\boldsymbol J_\omega^{\mathsf T}\boldsymbol I_{\mathrm{COM}}\boldsymbol J_\omega$
and corresponding velocity-dependent inertial force to $\boldsymbol h$.
The chosen mass representation supplies these terms once, with the
corresponding constant-mass approximation and duplicate eccentric-inertia
terms removed. The same equations close the load path and determine
motion: support deformation changes gear engagement, redistributed gear load
changes bearing and pin reactions, and those reactions return through the
structural and inertial coordinates.

\section{Implicit solution and accepted-step observations}
\label{sec:solution}

The implicit step closes the interaction between body motion and contact
reaction. A trial motion changes the local gaps and structural port
displacements; the resulting forces enter a common residual whose solution
updates that motion. This nonlinear closure is distinct from advancing
contact history or accumulating service exposure. The state-update
procedure below keeps those operations tied to the accepted trajectory.

\subsection{Time discretization and effective tangent}

The fast dynamics use the generalized-$\alpha$ method~\cite{m_chung1993}.
For a trial increment $\Delta t$, the acceleration at $t_{n+1}$ is the nonlinear
unknown. Define
\begin{align}
 \boldsymbol q_p&=\boldsymbol q_n+\Delta t\boldsymbol v_n
                 +\Delta t^2(1/2-\beta)\boldsymbol a_n, &
 \boldsymbol v_p&=\boldsymbol v_n+\Delta t(1-\gamma)\boldsymbol a_n,
 \label{eq:predictor}\\
 \boldsymbol q_{n+1}&=\boldsymbol q_p+\beta\Delta t^2\boldsymbol a_{n+1}, &
 \boldsymbol v_{n+1}&=\boldsymbol v_p+\gamma\Delta t\boldsymbol a_{n+1}.
 \label{eq:corrector}
\end{align}
The evaluation states are
\begin{align}
 \boldsymbol q_\alpha&=(1-\alpha_f)\boldsymbol q_{n+1}
                             +\alpha_f\boldsymbol q_n,\qquad
 \boldsymbol v_\alpha=(1-\alpha_f)\boldsymbol v_{n+1}
                             +\alpha_f\boldsymbol v_n,\nonumber\\
 \boldsymbol a_\alpha&=(1-\alpha_m)\boldsymbol a_{n+1}
                             +\alpha_m\boldsymbol a_n,
 \label{eq:alpha_states}
\end{align}
with
\begin{equation}
 \alpha_m=\frac{2\rho_\infty-1}{\rho_\infty+1},\quad
 \alpha_f=\frac{\rho_\infty}{\rho_\infty+1},\quad
 \gamma=\frac12+\alpha_f-\alpha_m,\quad
 \beta=\frac14(1-\alpha_m+\alpha_f)^2.
 \label{eq:alpha_parameters}
\end{equation}
The parameter $\rho_\infty$ controls algorithmic high-frequency dissipation;
it is distinct from physical contact and structural damping.

The residual is Eq.~\eqref{eq:eom} evaluated at these states, with the controls
and history input fixed for the current trial. Controls that depend on trial
duration are refreshed before retries. The operating protocol specifies their
time sampling; weighted kinematic states alone do not establish the order of an
arbitrarily sampled external load.
For constant mass, setting
$\boldsymbol K_q=-\partial\boldsymbol Q/\partial\boldsymbol q$ and
$\boldsymbol C_v=-\partial\boldsymbol Q/\partial\boldsymbol v$ gives
\begin{equation}
 \boldsymbol J_{\mathrm{eff}}
 =(1-\alpha_m)\boldsymbol M
 +(1-\alpha_f)\beta\Delta t^2\boldsymbol K_q
 +(1-\alpha_f)\gamma\Delta t\boldsymbol C_v,
 \qquad
 \boldsymbol J_{\mathrm{eff}}\Delta\boldsymbol a=-\boldsymbol r.
 \label{eq:effective_tangent}
\end{equation}
Configuration-dependent inertia introduces additional derivatives of
$\boldsymbol M\boldsymbol a_\alpha+\boldsymbol h$. In the supported native
mass-refresh path, the residual includes these inertial terms, whereas the
standard assembled tangent can freeze their derivatives. This is a modified
Newton treatment, not an exact linearization of the general equation.

The force tangent itself can combine exact linear blocks, local algorithmic
tangents, automatic differentiation with a fixed discrete contact candidate,
and geometric approximations. Candidate selection is outside the
differentiation tape; differentiability within a selected branch does not
imply differentiability across contact switching. A residual finite-difference
route is also available. The selected tangent, refresh frequency, and linear
solver are consequently reported as numerical settings. Nonsymmetric systems
are solved without assuming positive definiteness.

\subsection{History ownership, retries, and acceptance}

Contact evaluation may overwrite geometry caches, iteration variables,
physical history, and force snapshots. These arrays are registered according
to their role. Before attempting a step, the solver saves the mechanical state
and all registered history and iteration inputs. Every Newton or line-search
evaluation restores that common input state before applying the contact law.
In abstract form,
\begin{equation}
 (\boldsymbol Q^{(i)},\boldsymbol z_{\mathrm{trial}}^{(i)})
 =\mathcal E(\boldsymbol q_\alpha^{(i)},\boldsymbol v_\alpha^{(i)};
                    \boldsymbol z_n,\boldsymbol\lambda_n).
 \label{eq:trial_state}
\end{equation}
The evaluation index $i$ is not a physical time index. Repeating a trial must
therefore not integrate friction history repeatedly. Tangent evaluations use
the same history input and restore their temporary changes.

A strict acceptance policy compares the residual norm with
\begin{equation}
 \|\boldsymbol W\boldsymbol r\|_2
 \leq\max\!\left(\varepsilon_{\mathrm{abs}},
       \varepsilon_{\mathrm{rel}}\|\boldsymbol W\boldsymbol r_0\|_2\right),
 \label{eq:acceptance}
\end{equation}
where $\boldsymbol W$ is the declared scaling. The supported rotational scaling
converts moment residuals using a reference length, permitting comparison with
force residuals; it is optional and must not be assumed for historical runs.
On rejection, mechanical and registered history states are restored before a
smaller trial or termination. Step records distinguish convergence-accepted
advances from any configured forced commits, so the quality of a saved
trajectory can be assessed independently of whether its run completed.

After acceptance, the solver restores the step input history and evaluates the
final accepted $\alpha$ state once to commit the contact history. This avoids
taking history from the last line-search trial, which may have been rejected.
The physical clock advances by the accepted increment
$\Delta t_n^{\mathrm{acc}}$, even if the adaptive scheduler has already selected
a different increment for the next step.

\begin{figure}[pos=tbp]
\centering
\resizebox{\linewidth}{!}{\begin{tikzpicture}[x=1cm,y=1cm,
 box/.style={draw=ink!70,rounded corners=2pt,fill=pale,align=center,
   minimum height=1.15cm,text width=3.4cm,inner sep=6pt,outer sep=0pt,
   font=\small,
   execute at begin node={\hyphenpenalty=10000\exhyphenpenalty=10000\relax}},
 flow/.style={-{Stealth[length=2mm]},draw=ink,line width=.7pt},
 note/.style={font=\scriptsize,fill=white,inner sep=2pt,align=center}]

\node[box] (old) at (2,5)
 {Step input\\$q_n,v_n,a_n,z_n$};
\node[box] (trial) at (7,5)
 {Trial evaluation\\$q_\alpha^{(i)},v_\alpha^{(i)},r^{(i)}$};
\node[box] (tangent) at (12,5)
 {Tangent and update\\$J_{\rm eff}\Delta a=-r$};
\node[box] (accept) at (7,2.5)
 {Acceptance check\\\scriptsize Residual and finite state};
\node[box] (commit) at (12,2.5)
 {Commit history\\Advance by $\Delta t_{\rm acc}$};
\node[box,text width=5.2cm,fill=white] (sum) at (3.6,0)
 {Record / accumulate / update\\\scriptsize Each accepted duration is consumed once};
\node[box,text width=5.2cm] (replay) at (10.6,0)
 {Observation transaction\\\scriptsize Replay at accepted $\alpha$ state; restore};

\draw[flow] (old.east)--(trial.west);
\draw[flow] (trial.east)--(tangent.west);
\draw[flow,accent] (tangent.north)--(12,6.3)--(7,6.3)--(trial.north);
\node[note,text=accent,anchor=south] at (9.5,6.4)
 {iterate from the same history input};
\draw[flow] (trial.south)--(accept.north);
\draw[flow,dashed,accent] (accept.west)--(2,2.5)--(old.south);
\node[note,text=accent,anchor=north] at (3.5,2.3) {reject / retry};
\draw[flow] (accept.east)--(commit.west)
 node[midway,above=3pt,font=\scriptsize] {accept};
\draw[flow] (commit.south)--(12,1.2)--(10.6,1.2)--(replay.north);
\draw[flow] (replay.west)--(sum.east);
\end{tikzpicture}
}
\caption{State ownership during an implicit step. Nonlinear evaluations share
the step input history. A rejected trial cannot advance that history or the
physical clock. Observations replay the accepted weighted state in a separate
transaction before scheduled updates. The exact acceptance and update
configuration remains part of each run specification.}
\label{fig:accepted_state}
\end{figure}

\subsection{Observation and computational execution}

Registered observers replay the accepted $\alpha$ state inside an independent
transaction, consume its force snapshots, and restore temporary changes before
structural or other slow updates. Thus force observations correspond to the
accepted weighted state; they should not be relabelled as endpoint forces.
Each recorded force retains its interface identity, position, and evaluation
state. Integral observations are weighted by the actual accepted duration;
sampled maxima retain a different update rule. This prevents repeated
Newton evaluations or a changed next-step size from altering the physical
exposure assigned to a contact. Section~\ref{sec:observables} defines the
load, slip, and higher-moment statistics required by the consuming processes.
Where a sliding-speed snapshot is unavailable, only the supported load
statistics are accumulated. Long-window accumulators use double precision
even when contact kernels use single precision.

GPU execution batches independent local contacts and reduces their generalized
force contributions. Eligible solver paths capture repeated kernels in CUDA
graphs; compatible dense or sparse direct solves are selected according to the
problem and device. Host contact and configuration-dependent mass-refresh paths
have different synchronization costs. The execution specification therefore
records the contact and accumulation precision, tangent path, linear solver,
capture eligibility, and observation schedule together with the physical
device. These choices determine how the same mechanical problem is executed;
they leave its contact identities and force-and-motion mappings explicit.

\section{Transmission characteristics and slow-state feedback}
\label{sec:observables}

The assembled solution connects external transmission behaviour to internal
load transfer. Output motion retains the effect of every active contact and
support, while the local record preserves the forces, positions, slip, and
history needed to examine their individual consequences. This correspondence
is the basis for using one mechanical model to study precision, stiffness,
vibration, and durability. Each characteristic requires its own loading and
observation protocol; heat, lubrication, and material evolution additionally
require constitutive inputs beyond the mechanical record.

Two types of connection follow. A downstream observation, such as stress
recovery from a prescribed contact load, leaves the mechanical trajectory
unchanged. A feedback process updates a physical gap or friction parameter
and thereby changes the next mechanical solution. The configured slow
variables $\boldsymbol\zeta$ enter the held step controls
$\boldsymbol\lambda$ through their geometry or constitutive maps;
the contact-memory state $\boldsymbol z$ remains separately identified.
Let $\boldsymbol R$ denote the static residual under fixed external
loading and constraints, and let $\boldsymbol F_n$ collect the individual
normal contact loads. On a specified equilibrium and history branch,
linearization with respect to a slow-state change gives
\begin{equation}
 \boldsymbol{R}_{q}\Delta\boldsymbol{q}
 =-\boldsymbol{R}_{\zeta}\Delta\boldsymbol{\zeta},
 \qquad
 \Delta\boldsymbol{F}_n=
 \boldsymbol{F}_{n,q}\Delta\boldsymbol{q}
 +\boldsymbol{F}_{n,\zeta}\Delta\boldsymbol{\zeta}.
 \label{eq:obs_coupling_path}
\end{equation}
An altered local gap changes both its own load and the surrounding assembly
response. Re-equilibration redistributes load through the connected bearings
and structure; a new contact set requires the nonlinear solution.
Equation~\eqref{eq:obs_coupling_path} is used here to interpret feedback
locally on a specified equilibrium branch.

\subsection{A common mechanical model with different loading protocols}

A protocol specifies boundary conditions, drive and load histories, settling,
recording coordinates, and the observation interval. Reusing the model means
preserving its geometry, material assignments, contact identities, and
structural representation across protocols. Table~\ref{tab:performance_protocols}
summarizes the resulting division between mechanical information, additional
physical assumptions, and feedback. For comparisons between two slow states,
the mounting, shaft constraints, load schedule, and reference phase remain
the same, so that a change in the measured characteristic can be attributed
to the changed state.

\begin{table}[pos=tbp]
\caption{Performance protocols derived from the common mechanical assembly.
Each row specifies a separate observation or evolution problem; the rows do
not imply simultaneous operation of all processes in one calculation.}
\label{tab:performance_protocols}
\centering
\small
\begin{tabularx}{\linewidth}{@{}>{\raggedright\arraybackslash}p{0.14\linewidth}YY@{}}
\toprule
Characteristic & Protocol and retained mechanical information & Additional inputs and connection to mechanics \\
\midrule
Precision & Rotation at a declared speed and load; unwrapped input and output angles over a stated phase interval. & Signed nominal ratio and angular reference; trajectory observation. \\
\addlinespace[2pt]
Stiffness & Fixed input and mounting; output torque sequence, deflection, and loading branch. & Settling and tangent, secant, or fitting interval; equilibrium observation. \\
\addlinespace[2pt]
Loads and stress & Contact identity, force, position, and geometry at a selected state or over a load history. & Elastic properties and local pressure discretization; stress recovery is downstream. \\
\addlinespace[2pt]
Vibration & Driven motion; response coordinate, direction, timestamps, and reference-shaft phase. & Excitation, damping, retained structural bandwidth, and spectral window; trajectory observation. \\
\addlinespace[2pt]
Heat and friction & Accepted contact-load and slip statistics over representative operating windows. & Thermal network, loss and lubricant laws; configured temperature-to-gap and friction updates. \\
\addlinespace[2pt]
Wear and fatigue & Element-wise sliding exposure or load moments with declared cycle and service times. & Wear areas and coefficients, or calibrated life laws; wear updates gaps, while damage requires a defect adapter to alter mechanics. \\
\bottomrule
\end{tabularx}
\end{table}

For a rotating protocol, the output-referred transmission error is
\begin{equation}
 e_{\mathrm{TE}}(t)=
 [\theta_{\mathrm{out}}(t)-\theta_{\mathrm{out},0}]
 -\frac{\theta_{\mathrm{in}}(t)-\theta_{\mathrm{in},0}}{i},
 \qquad i=\frac{\omega_{\mathrm{in}}}{\omega_{\mathrm{out}}},
 \label{eq:obs_te}
\end{equation}
where $i$ is the signed ratio determined by the transmission topology.
Angles are continuously unwrapped. The implemented transmission-error
analysis retains this nominal ratio; a fitted ratio is diagnostic and cannot
silently remove physical drift. The displayed trace may remove a constant
mean, with its offset reported separately. Peak-to-peak and RMS values belong to a
declared time or shaft-angle interval. A partial rotation does not establish
full-output-period accuracy without an additional closure argument.

For stiffness, the input and housing constraints are prescribed and the
output load is varied. Torque and output deflection define a branch-dependent
relation $T(\theta;\boldsymbol{\zeta})$, where $\boldsymbol{\zeta}$ denotes
the frozen slow state. Its local tangent is $K_t=\partial T/\partial\theta$;
a secant or interval fit is a separate estimator. The role of internal
deformation follows directly from the static residual. Partition the free
coordinates into output rotation $\theta$ and the remaining coordinates
$\boldsymbol r$. Let $\boldsymbol R^s(\theta,\boldsymbol r;
\boldsymbol\zeta)$ contain the restoring generalized forces minus all
prescribed loads except the independently applied output torque, so that
equilibrium requires $\boldsymbol R^s=(T,\boldsymbol 0)^{\mathsf T}$.
On a differentiable branch with a specified contact history, and with the
input and slow state fixed,
\begin{equation}
 \begin{bmatrix}
 K^s_{\theta\theta}&\boldsymbol K^s_{\theta r}\\
 \boldsymbol K^s_{r\theta}&\boldsymbol K^s_{rr}
 \end{bmatrix}
 \begin{bmatrix}\Delta\theta\\\Delta\boldsymbol r\end{bmatrix}
 =\begin{bmatrix}\Delta T\\\boldsymbol 0\end{bmatrix},
 \qquad
 K_t=K^s_{\theta\theta}
 -\boldsymbol K^s_{\theta r}
 (\boldsymbol K^s_{rr})^{-1}\boldsymbol K^s_{r\theta},
 \label{eq:obs_condensed_stiffness}
\end{equation}
where $\boldsymbol K^s=\partial\boldsymbol R^s/\partial(\theta,
\boldsymbol r)$ and $\boldsymbol K^s_{rr}$ is assumed nonsingular. The
second term accounts for relaxation of the internal coordinates through the
contacts, bearings, and retained structural deformation. It explains why
assembly stiffness generally differs from a stiffness computed with all
internal centres and orientations fixed. The condensation is an
interpretation of the equilibrium equations, identifying the internal load
paths that contribute to the output response. The application below uses
branch fits rather than a numerical evaluation of this condensed tangent.

At fixed input rotation, $\partial e_{\mathrm{TE}}/\partial T=1/K_t$ on
the same branch. Precision under load and torsional stiffness thus observe
the same equilibrium response. Contact opening, load reversal, and slip
transitions can change that branch; an interval fit across several load
levels need not equal its local tangent. Complete signed loading sequences
are retained, and $\oint T\,\mathrm{d}\theta$ is interpreted as cycle
dissipation only after closure and repeatability have been established.

Vibration adds a time and observation-coordinate requirement. The body,
direction, and retained boundary coordinate define the measured response;
an output torsional signal and a housing acceleration sample different
transmission paths. For time spectra, the accepted timestamps determine
whether uniform-time resampling is required; shaft-order spectra use the
declared unwrapped phase. For $N_r$ observed reference revolutions, the order spacing
is $1/N_r$. A meaningful frequency range also requires the corresponding
structural modes and support dynamics. Nominal mesh and bearing frequencies
guide interpretation, but a frozen cage cannot generate rolling-passage
excitation, and ideal symmetry can suppress excitation at a selected housing
coordinate.

\subsection{Contact histories and process-specific exposure}

Let $F_{c,n}$ and $v_{s,c,n}$ be the normal force and slip speed of element
$c$ at the accepted evaluation state, and let $\Delta t_n^a$ be the actual
accepted duration. Over a sampling window of duration $\tau_w$, the relevant
statistics include
\begin{equation}
 \begin{aligned}
 I_{F,c}&=\sum_n F_{c,n}\Delta t_n^a,&
 I_{s,c}&=\sum_n F_{c,n}|v_{s,c,n}|\Delta t_n^a,\\
 I_{m,c}&=\sum_n F_{c,n}^{m}\Delta t_n^a,&
 F_{\mathrm{pk},c}&=\max_n F_{c,n}.
 \end{aligned}
 \label{eq:obs_statistics}
\end{equation}
Repeated residual evaluations do not provide additional physical exposure.
Integral quantities and sampled maxima therefore have different update rules.
The implementation can additionally retain load-weighted contact positions
and normals, which support spatial thermal-source and deformation mappings.
The element identity is preserved: summing all forces would discard the
load concentration that controls local pressure, wear, and fatigue.

The window itself belongs to the consuming process. A mesh-period window
can be suitable for mean tooth-contact power, whereas a rolling-element
process may require a complete cage or race-relative cycle. Phase is formed
from declared combinations of shaft coordinates. The current cycle-window
implementation apportions the final accepted-step integral at the phase
boundary; the sampled peak remains a maximum. This removes an arbitrary
fixed-step window boundary, but does not replace time-step convergence or
resolve every slower modulation. These statistics form a sufficient input
only for the declared consumer: a mean force cannot replace a fatigue load
moment, and a force maximum cannot recover the sliding exposure needed for
wear or heat.

\subsection{Distributed loads and local stress recovery}

The mechanical contact solution supplies element loads and the geometry at
which they act. For the finite-width pressure model, transverse panels span
the face width $b$, with $f_j=p_jb\Delta x_j$. A two-body compliance matrix
$\boldsymbol{A}$ is obtained by integrating the Boussinesq $1/r$ influence
over each rectangular panel. Pressure and approach $d$ satisfy
\begin{equation}
 \boldsymbol{g}=\boldsymbol{g}^{0}
       +\boldsymbol{A}\boldsymbol{f}-d\boldsymbol{1}\geq\boldsymbol{0},
 \quad \boldsymbol{f}\geq\boldsymbol{0},\quad
 f_jg_j=0,\quad \sum_j f_j=F_n .
\label{eq:obs_pressure_contact}
\end{equation}
For a receiver at the centre of panel $i$, the coefficient is
\begin{equation}
 A_{ij}=\frac{1}{\pi E^* b\Delta x_j}
 \int_{x_j^-}^{x_j^+}\int_{-b/2}^{b/2}
 \frac{\mathrm{d}y\,\mathrm{d}\xi}
 {\sqrt{(x_i-\xi)^2+y^2}},\qquad
 \frac{1}{E^*}=\frac{1-\nu_a^2}{E_a}+\frac{1-\nu_b^2}{E_b}.
 \label{eq:obs_panel_influence}
\end{equation}
The rectangular integral, including the integrable self-panel singularity,
is evaluated analytically. The factor $b\Delta x_j$ converts a force
unknown into the applied panel traction. The undeformed gap is sampled
from the final modified wheel profile and the circular pin at the selected
mechanical pose. A constant gap datum is absorbed by the unknown approach;
changing that datum does not change the prescribed-force pressure problem.
Thus the pressure unknowns are distributed in one transverse direction,
while the compliance includes the finite face-width integration. This
calculation uses the selected contact load; it does not automatically replace
the contact law used in the global dynamic residual.

The stored pressure is then interpolated along the surface coordinate and
transformed into a line-contact half-space stress field. Fourier-domain
transfer functions recover $\boldsymbol{\sigma}(x,z)$ from the prescribed
normal and tangential tractions. This recovery assumes invariance along the
face-width direction, so three-dimensional face-edge stress variations are
outside that approximation. The reported maximum shear stress is
$(\lambda_{\max}(\boldsymbol{\sigma})-
\lambda_{\min}(\boldsymbol{\sigma}))/2$.
Raw panel maxima and finite-width spatial averages are distinct pressure
observables. Their ranking can differ from the ranking of total element
forces, and each requires its own spatial-resolution assessment.
The assessment holds the mechanical state, contact identity, force,
materials, profile representation, and physical panel domain fixed.
Increasing the number of panels in that domain then tests local pressure
resolution independently of the global time step or structural reduction.

\subsection{Slow feedback through shared mechanical channels}

The accepted contact statistics also provide inputs to thermal, lubrication,
wear, and fatigue protocols. The sliding exposure $I_s$ is not itself
frictional heat: the estimate $\overline P_c=\mu_c I_{s,c}/\tau_w$ requires
a declared sliding-loss model, and a tangential law with recoverable storage
requires a compatible dissipation definition. A thermal network converts
the assigned source vector into temperatures; configured expansion maps
then update bearing clearances or contact gaps. Lubrication can use the
temperature, load, and surface inputs to update the friction coefficient
for the next mechanical window. Temperature-dependent modulus in the film
calculation does not by itself update the structural stiffness.

Wear consumes sliding exposure with a specified material coefficient and
contact-to-surface association, then writes material loss to the wear
contribution of the contact gap. The single-channel protocol limits the
geometry increment and scales the represented service time consistently;
mechanical settling after the update is excluded from the next exposure
window. Fatigue instead consumes a calibrated load moment or stress
spectrum with a declared cycle rate. Its accumulated damage is a downstream
state unless a spatial defect adapter is connected to the mechanical model.

These updates act on the contact geometry or constitutive parameters
already used by the assembly. Re-solving the mechanical problem is what
turns a local thermal or wear increment into changes of load distribution,
transmission error, stiffness, or vibration, as expressed locally by
Eq.~\eqref{eq:obs_coupling_path}. Comparisons between successive slow states
therefore repeat the relevant performance protocol under matched loading,
support conditions, and observation phase.

The implemented processes have distinct update schedules and material
inputs. Combining them requires compatible state updates, common
service-time accounting where appropriate, and the required coefficient
refreshes. Their presence does not establish a simultaneous, generally
validated multiprocess solver. Appendix~\ref{sec:supp_slow_processes}
gives the thermal, lubrication, wear, and fatigue formulations, a minimal
thermal feedback construction, and their composition boundaries.
The numerical examples in this article examine the mechanical assembly,
structural reduction, load transfer, and local pressure; they do not
calibrate or validate thermal or durability predictions.

\section{Verification of contact calculations and time integration}
\label{sec:verification}

Verification is organized at the level at which a reference can be defined.
The tests in this section were executed with the production contact functions
and implicit solver. They address implementation error and smooth-problem time
discretization, respectively. The engineering examples in
Section~\ref{sec:applications} are evaluated separately; agreement in a contact
primitive is not used as an error bound for a complete reducer.

\subsection{Contact laws, signed clearance, and force direction}

The point-contact and linearized line-contact primitives are compared with
independent double-precision evaluations of
$F_n=k_p\langle-g\rangle_+^{3/2}$ and
$F_n=k_\ell\langle-g\rangle_+$.
The coefficients have the units implied by the corresponding exponent.
The reference evaluates the actual single-precision input values promoted to
double precision, so the comparison measures arithmetic and implementation
error without conflating it with initial parameter rounding.
For a reference force sequence, the normalized error is
\begin{equation}
 \epsilon_F=
 \frac{\max_j\|\boldsymbol F_j-\boldsymbol F_j^{\mathrm{ref}}\|_2}
      {\max_j\|\boldsymbol F_j^{\mathrm{ref}}\|_2}.
 \label{eq:force_error}
\end{equation}
For a scalar law, the vector norm reduces to the absolute value.

External and internal circle-contact configurations additionally test signed
clearance and force direction. Their reference gaps are
$g_{\rm ext}=\|\boldsymbol c_B-\boldsymbol c_A\|-r_A-r_B$ and
$g_{\rm int}=R_B-r_A-\|\boldsymbol c_B-\boldsymbol c_A\|$.
The reference force points away from the other external circle or toward the
centre of the enclosing circular surface. The test includes both separated
and compressed configurations over different centre directions.
The maximum absolute gap errors are $1.40\times10^{-9}$~m externally and
$6.55\times10^{-10}$~m internally.

\begin{table}[pos=tbp]
\caption{Production contact primitives versus independent double-precision
reference evaluations. The normalization in Eq.~\eqref{eq:force_error}
uses the peak reference force of each test.}
\label{tab:contact_validation}
\centering
\begin{tabular}{@{}lrr@{}}
\toprule
Test & Maximum force error (N) & Normalized error\\
\midrule
Point-contact law & $1.4554\times10^{-4}$ & $7.2770\times10^{-8}$\\
Linearized line-contact law & $3.0464\times10^{-5}$ & $3.0464\times10^{-8}$\\
External circle contact & $1.3702\times10^{-2}$ & $6.8512\times10^{-6}$\\
Internal circle contact & $6.4850\times10^{-3}$ & $3.2425\times10^{-6}$\\
\bottomrule
\end{tabular}
\end{table}

\begin{figure}[pos=tbp]
\centering
\includegraphics[width=\linewidth]{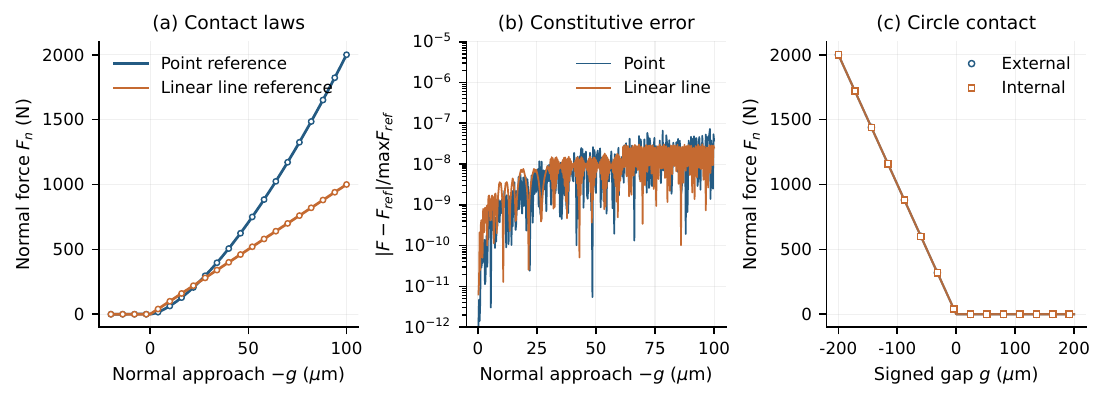}
\caption{Component verification. The signed approach shown in the constitutive
tests is $-g$; only its positive part is the physical indentation.
The GPU functions reproduce the independent point and linearized line-contact
relations. Separate internal and external circle tests check the vector force
and its direction. These checks do not test the discretized cycloidal search
or a finite-width gear contact model.}
\label{fig:contact_validation}
\end{figure}

Table~\ref{tab:contact_validation} and \figref{fig:contact_validation} show
agreement within the declared single-precision error budgets. Separated
configurations have zero force; the separation check uses reference gaps
above $10^{-7}$~m to exclude ambiguity at the rounding-scale contact boundary.
The nonzero normal forces have the expected compressive direction.
These results verify the selected constitutive and geometric primitives,
including their sign conventions.

\subsection{Temporal convergence of the production implicit solver}

A one-coordinate oscillator isolates the time-integration error:
\begin{equation}
 m\ddot q+kq=0,\qquad
 m=1~\mathrm{kg},\quad
 k=(2\pi\,50)^2~\mathrm{N/m},\quad
 q(0)=0.01~\mathrm m,\quad \dot q(0)=0.
 \label{eq:oscillator}
\end{equation}
The initial acceleration is set consistently to $-kq(0)/m$.
The reference is $q_{\rm ref}(t)=q(0)\cos(\sqrt{k/m}\,t)$, evaluated using the
coefficient values passed to the solver. One nominal period, $T=0.02$~s,
is integrated with $N=20,40,80,160,320$ uniform increments.
The relative displacement error and observed order are
\begin{equation}
 \epsilon_q(N)=
 \frac{\max_j|q_j-q_{\rm ref}(t_j)|}{|q(0)|},\qquad
 p(N)=\log_2\frac{\epsilon_q(N)}{\epsilon_q(2N)}.
 \label{eq:time_error}
\end{equation}

The production solver uses $\rho_\infty=0.8$, an exact scalar spring tangent,
up to four Newton iterations, and strict step acceptance with an absolute
force-residual threshold of $10^{-3}$~N. The state and residual are stored
in single precision and the selected one-block linear solve uses double
precision. This is a small-system path of the production solver, not a
large sparse-solver benchmark.

\begin{figure}[pos=tbp]
\centering
\includegraphics[width=\linewidth]{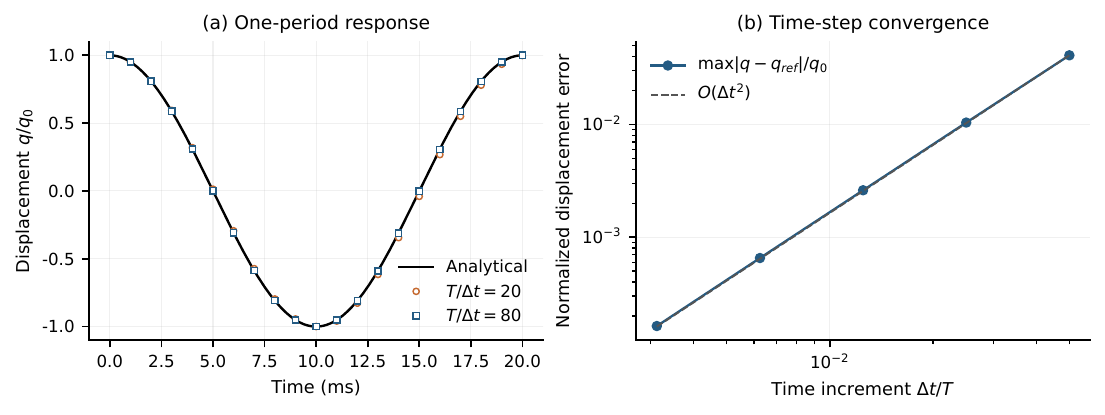}
\caption{Time-integration verification using the production generalized-$\alpha$
solver. (a) Numerical displacement and the independent analytical oscillator
solution. (b) Maximum displacement error over one period under uniform
time-step refinement. The experiment uses consistent initial acceleration;
the observed slopes concern this smooth problem.}
\label{fig:temporal}
\end{figure}

Refining the increment from $T/20$ to $T/320$ reduces $\epsilon_q$ from
$4.1151\times10^{-2}$ to $1.6260\times10^{-4}$.
The consecutive orders are 1.984, 1.996, 1.998, and 2.005
(\figref{fig:temporal}). All accepted times sum to one period, with no
unconverged commits, graph fallbacks, or slow retries. The largest recorded
residual is $3.052\times10^{-5}$~N, below the prescribed threshold.
The tests were run on physical GPU~0, an NVIDIA GeForce RTX~3070~Ti Laptop GPU
with 8~GiB memory, using Warp~1.16.0. The stated inputs, method,
acceptance condition, and error definition specify this idealized test.

The approximately second-order result establishes the smooth-problem temporal
behaviour of the selected solver path. Contact switching, friction transitions,
profile interpolation, and delayed structural updates introduce additional
errors that require dedicated convergence studies. The recorded oscillator
wall times include setup and host observations and are not used to claim
GPU acceleration or reducer-level throughput.

\section{Structural reduction at distributed contact ports}
\label{sec:ports_example}

\subsection{Component model and comparison scope}

Structural reduction must preserve the displacement caused by contact
loading both at the loaded location and elsewhere in the assembly.
The question examined here is whether a compact representation with an
accurate static compliance also preserves dynamic transmission between
contact ports. The three-dimensional annular housing in
Table~\ref{tab:ports_inputs} provides a fixed component model for this
comparison. Its continuous geometry and linear mounting springs are
idealizations; the material and mounting values are prescribed inputs.
The model is constructed with the framework's finite-element and
port-reduction functions described in Section~\ref{sec:structural_ports}.

\begin{table}[pos=tbp]
\caption{Annular housing model and discretization. Mounting and material
values are prescribed for this component comparison.}
\label{tab:ports_inputs}
\centering\small
\begin{tabularx}{\linewidth}{@{}p{.39\linewidth}Y@{}}
\toprule
Quantity & Value\\
\midrule
Inner / outer radius; axial height & 105 / 125 mm; 67.6 mm\\
Young's modulus / Poisson ratio / density & 210 GPa / 0.3 / 7850 kg/m$^3$\\
Contact locations & 52 nodal ports, each with three translations\\
Mounting & 18 nodal spring locations in a lower-face neighbourhood;
nominal radius 116.5 mm\\
Mount stiffness & $5.556\times10^6$ N/m per location and Cartesian direction\\
Finite elements & First-order displacement tetrahedra; nominal mesh size 6.667 mm\\
Generated mesh & 3638 vertices, 11864 elements, 10914 displacement DOFs\\
Parent Craig--Bampton model & 156 boundary components and 20 fixed-interface modes\\
Damping & Rayleigh calibration with $\zeta=0.02$ at the constructor's
calibration frequencies; no added mount damping\\
Frequency-response sampling & 10--5000 Hz, increment 2.5 Hz\\
\bottomrule
\end{tabularx}
\end{table}

The finite-element model is first reduced to 176 Craig--Bampton
coordinates. All subsequent representations use these same parent
matrices, boundary groups, and damping. Consequently, \emph{full basis}
denotes all modes of this 176-coordinate parent. The comparison isolates
the second reduction to structural coordinates and contact-port maps;
finite-element mesh convergence and truncation from the original
10914-coordinate model to the parent are not assessed here.

\begin{figure}[pos=tbp]
\centering
\includegraphics[width=\linewidth]{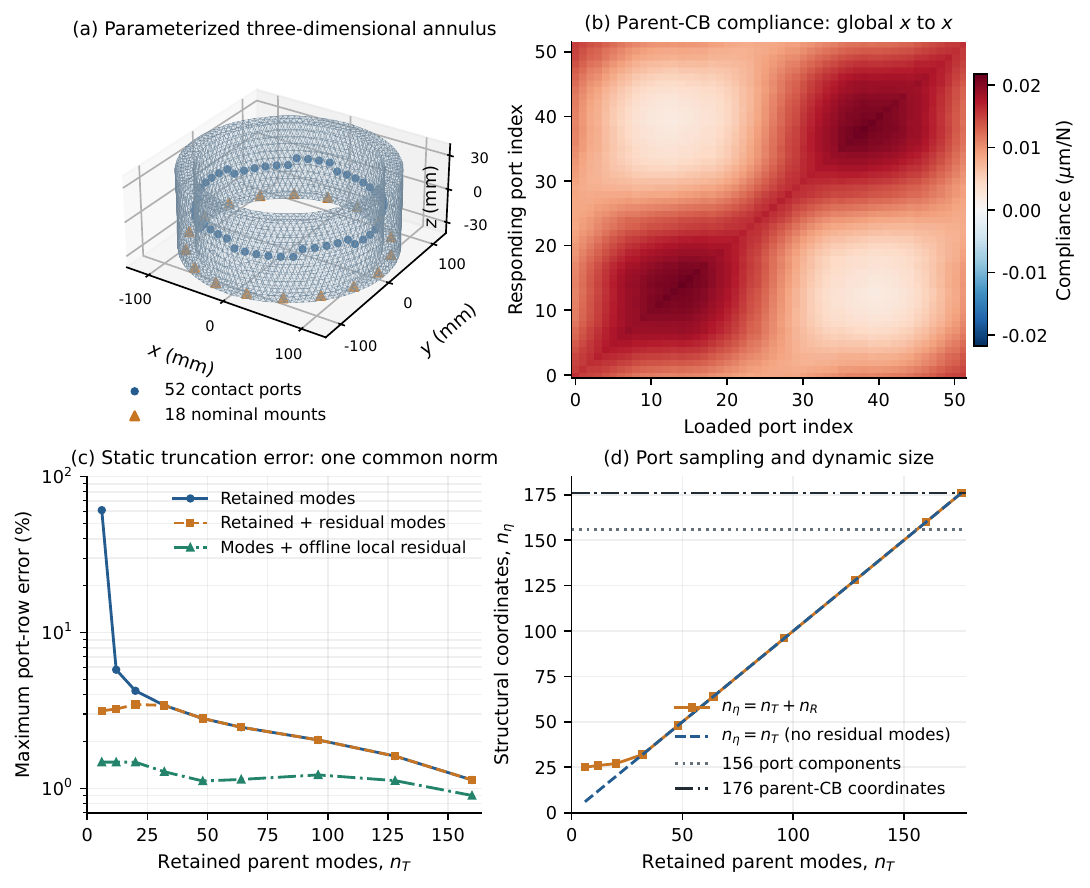}
\caption{Structural information retained at the contact ports.
(a) Generated annulus surface mesh with contact and nominal mounting locations.
(b) Parent-model compliance between global $x$ components, not between
separately rotating radial directions.
(c) Static port-row error under different reductions.
(d) Structural coordinate counts compared with the fixed boundary dimension.
The separately labelled local-residual correction is an offline reconstruction;
it is not included in the present contact runtime. The full-basis identity
is reported numerically in the text and excluded from the truncated-error plot.}
\label{fig:ports_static}
\end{figure}

\subsection{Consistency of the structural interface}

The reference static operator is
$\boldsymbol C^{\rm ref}=\boldsymbol S_b\boldsymbol K_r^{-1}
\boldsymbol S_b^{\mathsf T}$.
Using all 176 generalized modes gives a relative Frobenius difference of
$1.22\times10^{-11}$. Direct complex solves of the parent equations at
100, 500, 1500, 3000, and 5000 Hz agree with the complete modal response
to a maximum relative difference of $2.73\times10^{-11}$.
These checks use double-precision CPU calculations, consistent with the
offline structural construction.

A separate deterministic linear-contact trace test exercises the complete
map in Eq.~\eqref{eq:port_runtime_transform}, including mixed rigid--port
terms. Virtual work agrees to $2.07\times10^{-14}$ at its specified
perturbation. Contracting the extended tangent and first transforming
the trace vectors before assembly agree to $1.37\times10^{-16}$.
They verify work and tangent consistency of the linear maps, including
the interaction of rigid and structural coordinates. Nonlinear contact
branches and their GPU implementation require separate checks.

The same 12-mode basis produces maps of sizes
$39\times12$, $78\times12$, and $156\times12$ when 13, 26, or 52 existing
locations are sampled. The number of structural unknowns remains twelve.
Thus, resolving additional contact locations does not necessarily enlarge
the dynamic state. It does change the spatial load representation, whose
resolution remains a separate choice from the retained structural basis.

\subsection{Static compliance and dynamic load transmission}

The static comparison resolves the error at each port through
\begin{equation}
 \epsilon_C=\max_g
 \frac{\|(\boldsymbol C-\boldsymbol C^{\rm ref})_{g,:}\|_{\rm F}}
      {\|\boldsymbol C^{\rm ref}_{g,g}\|_{\rm F}},
 \label{eq:ports_static_error}
\end{equation}
where $g$ selects the three Cartesian rows of a single port.
The numerator includes its complete $3\times156$ row block and the
denominator is the reference $3\times3$ self-block.

For unit load direction $\boldsymbol d$ and measured direction
$\boldsymbol s$, the reduced compliance frequency response is
\begin{equation}
 G_{sd}(\omega)=
 \boldsymbol s^{\mathsf T}\boldsymbol H
 \left[\boldsymbol K_\eta-\omega^2\boldsymbol I+
       {\rm i}\omega\boldsymbol C_\eta\right]^{-1}
 \boldsymbol H^{\mathsf T}\boldsymbol d.
 \label{eq:ports_frf}
\end{equation}
The input is global $x$ at port~0, its nominal radial direction.
The receivers are that same direction and the nominal radial direction
at port~8. Table~\ref{tab:ports_errors} reports
$\epsilon_G=\|\boldsymbol G-\boldsymbol G^{\rm ref}\|_2/
\|\boldsymbol G^{\rm ref}\|_2$ over the sampled complex-valued response.
The reference is the full-basis response checked against the parent solves.

\begin{table}[pos=tbp]
\caption{Errors relative to the same parent CB model. The offline local
correction adds a block-diagonal residual compliance outside the dynamic coordinates.}
\label{tab:ports_errors}
\centering\small
\begin{tabular}{@{}lrrrr@{}}
\toprule
Representation & Coordinates & $\epsilon_C$ (\%) &
Driving $\epsilon_G$ (\%) & Transfer $\epsilon_G$ (\%)\\
\midrule
12 retained modes & 12 & 5.780 & 10.388 & 0.679\\
6 retained + 19 residual modes & 25 & 3.151 & 14.421 & 24.975\\
12 retained + 14 residual modes & 26 & 3.234 & 5.103 & 0.312\\
\shortstack[l]{26-coordinate model\\+ offline local correction} & 26 & 1.477 & 0.141 & 0.312\\
\bottomrule
\end{tabular}
\end{table}

\begin{figure}[pos=tbp]
\centering
\includegraphics[width=\linewidth]{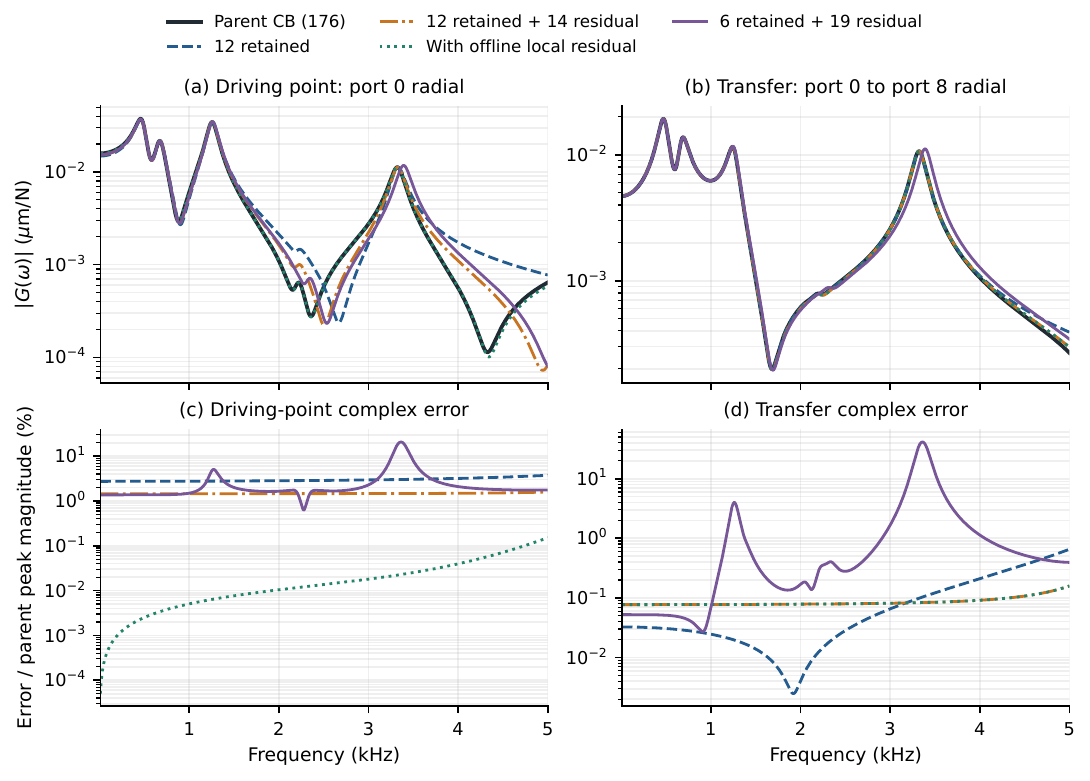}
\caption{Driving-point and cross-port frequency responses of the fixed
structural example. Upper panels show magnitudes; lower panels show
complex-response differences divided by the corresponding parent response's
peak magnitude over the band. This avoids division by an antiresonance zero.
The 25-coordinate and 26-coordinate models have similar static port-row
errors but different dynamic transfer errors.
The local-residual curve is an offline diagnostic.}
\label{fig:ports_frf}
\end{figure}

The main distinction appears between the 25-coordinate and
26-coordinate representations. Their static port-row errors are nearly
equal, yet their transfer errors are 24.975\% and 0.312\%.
Six retained parent modes extend to about 678 Hz; twelve extend to
about 3351 Hz. Residual attachment modes restore part of the compliance
but do not replace each omitted parent mode and its resonance.
Selecting the reduced basis therefore requires the frequency band and
the locations of loading and observation, in addition to a target
coordinate count or static error.

For the 26-coordinate case, adding the exported local self-blocks reduces
the offline static error from 3.234\% to 1.477\%.
The current runtime does not consume this additional self-block correction,
so the latter value is not assigned to its mechanical solution.
The correction is frequency-independent and block-diagonal:
$\boldsymbol s_8^{\mathsf T}\boldsymbol R_{\rm local}\boldsymbol d_0=0$.
It cannot change the selected cross-port transfer, explaining the
coincident corrected and uncorrected transfer curves.

The structural port representation makes this distinction explicit:
contact locations define where loads enter and displacements are
evaluated, while the retained coordinates determine which structural
responses can be transmitted between them. The present component study
establishes that relationship for one parent model and frequency band.
It does not measure the effect of housing flexibility on the reducer
application below. The idealized inputs and comparison definitions
are specified in Appendix~\ref{app:ports_reproduction}.

\section{Load transfer through a shared output pin}
\label{sec:coupling_example}

\subsection{Prescribed motion and cross-compliance}

The shared-pin interface provides a second, distinct mechanism for
transmitting a local displacement through the assembly. In this case,
the question is whether moving one wheel changes the reaction on another
wheel whose pose remains fixed. The comparison varies the cross-station
compliance in Eq.~\eqref{eq:pin_compliance} while preserving the
self-compliance, geometry, and sleeve clearance. It uses the framework's
two-wheel output-pin force and directional-tangent functions, including
slice integration and reaction assembly. Prescribing the poses isolates
the local interface response; no whole-reducer equilibrium is solved.

The coordinate order is
\begin{equation}
 \boldsymbol q=(x_f,y_f,\theta_f,
                 x_0,y_0,\theta_0,x_1,y_1,\theta_1)^{\mathsf T},
 \label{eq:coupling_coordinates}
\end{equation}
where $f$ denotes the flange and each angle is about the $z$ axis.
Translations are measured in metres and rotations in radians. The
corresponding body-force triples are $(F_x,F_y,M_z)$, with moments taken
about the respective body origins. Axial translation and out-of-plane
rotation are constrained. Table~\ref{tab:coupling_inputs} specifies the
geometric and constitutive inputs. The compliance is prescribed as a
local model parameter, independently of the annular housing in the
preceding example and of the reducer application that follows.

\begin{table}[pos=tbp]
\caption{Inputs to the shared-pin example. Both cases retain identical
self-compliance, geometry, and sleeve clearance.}
\label{tab:coupling_inputs}
\centering\small
\begin{tabularx}{\linewidth}{@{}p{.43\linewidth}Y@{}}
\toprule
Quantity & Value or convention\\
\midrule
Pin and hole distribution radii & $R=35$~mm for both\\
Outer pin and wheel-hole radii & $1.50$ and $1.55$~mm\\
Whole-contact power-law coefficient & $k=2\times10^8~\mathrm{N}\,\mathrm{m}^{-10/9}$\\
Contact width and integration & $6$~mm; four equal-weight midpoint slices\\
Slice stations relative to each wheel & $s=\{-2.25,-0.75,0.75,2.25\}$~mm\\
Wheel midplanes relative to flange & $z_0=-3$~mm, $z_1=3$~mm\\
Self- and cross-compliance & $C_{00}=C_{11}=1.5\times10^{-9}$~m/N;
 $C_{01}=0$ or $0.9\times10^{-9}$~m/N\\
Sleeve model & Massless; $5~\mu$m radial clearance disk;
 inner Hertz indentation omitted\\
Prescribed translations & $x_0=100$--$120~\mu$m at 81 equally spaced
 points; $x_1=-70~\mu$m\\
Remaining inputs & One pin; zero phases, other coordinates, velocities,
 damping, and thermal offsets; no tangential friction\\
Numerical regularization & Sleeve-update tangent floor $10^3$~N/m;
 determinant floor $10^{-12}$; distance and open-tangent indentation
 floors $10^{-9}$~m\\
\bottomrule
\end{tabularx}
\end{table}

Deleting only $C_{01}$ defines the reference while preserving each station's
self-compliance. Both cases remain on the same sleeve-seating branch:
all eight slice indentations exceed $1~\mu$m, and the sleeve displacement
magnitude differs from the clearance radius by less than $2$~nm throughout
the 162-state comparison. The load changes can therefore be interpreted
on one seating branch, without conflating cross-compliance with a
transition in the sleeve-clearance condition.

\subsection{Local closure and transmission of wheel load}

At each trial sleeve position, the algorithm evaluates the open-contact
resultants and tangents in Eq.~\eqref{eq:pin_open_state}, solves
Eq.~\eqref{eq:pin_local_closure}, and clips tensile candidate forces.
Starting from $\boldsymbol u_{\mathrm{fit}}^0=\boldsymbol0$, the sleeve
centre undergoes 16 updates,
\begin{equation}
 \boldsymbol u_{\mathrm{fit}}^{k+1}=
 \Pi_{\|\boldsymbol u\|\leq c}
 \left(\boldsymbol u_{\mathrm{fit}}^k
       -0.7\frac{\sum_{i=0}^{1}F_i^c\boldsymbol n_i}
                    {\max(K_0+K_1,10^3~\mathrm{N/m})}\right),
 \qquad c=5~\mu\mathrm m.
 \label{eq:coupling_fit_update}
\end{equation}
Here $\boldsymbol n_i$ points in the force direction on wheel $i$,
so its opposite acts on the sleeve. Projection $\Pi$ leaves an interior
trial unchanged and radially rescales an exterior trial to radius $c$.
At the final position,
the algorithm repeats the local closure and reevaluates every nonlinear
slice force after subtracting the station deflection. These reevaluated
forces supply the wheel and flange reactions. The following results
characterize this finite-iteration, single-precision algorithm; they
are not identified with a fully converged nonlinear complementarity
solution.

The contact normals oppose each other in this scan, giving
$\boldsymbol n_0\cdot\boldsymbol n_1=-1$. Consequently, the projected
station correction is $u_1=C_{11}F_1^c-C_{01}F_0^c$. Increased wheel-0
loading reduces this correction and increases the effective indentation
at wheel~1. The coupling therefore depends on force direction as well as
the positive cross-compliance coefficient.

\begin{figure}[pos=tbp]
\centering
\includegraphics[width=\linewidth]{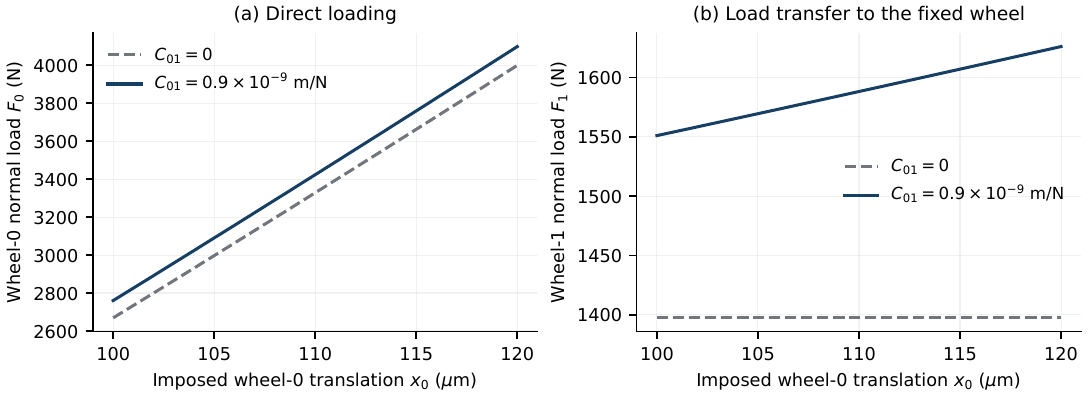}
\caption{Prescribed-motion response of the shared output pin.
(a) Wheel-0 loading as its translation increases.
(b) Wheel-1 load at its fixed pose, $x_1=-70~\mu$m.
The two 81-point scans differ only in $C_{01}$ and remain on the same
sleeve-seating branch. Parameters are given in
Table~\ref{tab:coupling_inputs}.}
\label{fig:coupling_response}
\end{figure}

As $x_0$ increases from 100 to $120~\mu$m, wheel-0 force rises from
2669.050 to 3999.531~N without cross-compliance and from 2760.192 to
4098.556~N with cross-compliance (\figref{fig:coupling_response}).
Wheel-1 force remains 1397.464~N in the reference but rises from
1550.975 to 1625.946~N in the coupled case despite its fixed pose.
At $x_0=110~\mu$m, the computed transfer derivative
$\partial F_1/\partial x_0$ is $2.209\times10^{-7}$ and
$3.750776$~N/$\mu$m, respectively. The reference retains the small
finite-precision value. The moving wheel changes the shared pin's
deformation and hence the other wheel's indentation. That dependence
would be lost if the two wheel loads were evaluated with independent,
unchanged supports.

\subsection{Reaction and tangent consistency}

\begin{figure}[pos=tbp]
\centering
\includegraphics[width=\linewidth]{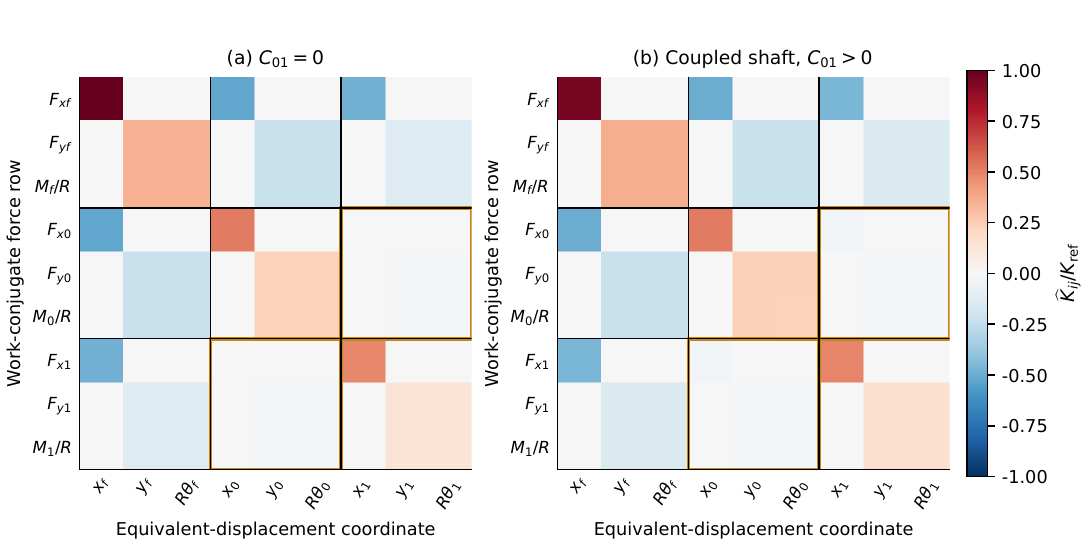}
\caption{Directional tangent at $x_0=110~\mu$m, expressed using
Eq.~\eqref{eq:coupling_tangent_scaling}. Both panels share
$K_{\mathrm{ref}}=\max_{i,j,\mathrm{case}}|\widehat K_{ij}|$.
Outlined blocks map one wheel's motion to the other wheel's reaction.
The matrices illustrate the implemented coupling structure. Independent
moment assembly checks the geometric force-arm contribution as described
in Appendix~\ref{app:moment_diagnostic}.}
\label{fig:coupling_tangent}
\end{figure}

For the solver-sign tangent $\boldsymbol K_q=-\partial\boldsymbol Q/
\partial\boldsymbol q$, a common unit is required before comparing blocks.
With $\boldsymbol S=\operatorname{diag}(1,1,1/R,1,1,1/R,1,1,1/R)$,
\begin{equation}
 \widehat{\boldsymbol q}=\boldsymbol S^{-1}\boldsymbol q,
 \qquad \widehat{\boldsymbol Q}=\boldsymbol S\boldsymbol Q,
 \qquad \widehat{\boldsymbol K}=\boldsymbol S\boldsymbol K_q\boldsymbol S.
 \label{eq:coupling_tangent_scaling}
\end{equation}
Thus, each coordinate triple is $(x,y,R\theta)$, each force triple is
$(F_x,F_y,M_z/R)$, and all tangent entries have units N/m.
Figure~\ref{fig:coupling_tangent} shows both scaled matrices and identifies
their inter-wheel blocks.
Some cross-wheel entries remain even when $C_{01}=0$, because the
clearance-constrained sleeve and flange are still shared. The comparison
removes the pin's cross-station compliance, not every interaction in the
assembly. The fixed-seating radial scan specifically isolates the change
in $\partial F_1/\partial x_0$ due to that compliance.

The maximum resultant-force imbalance in the scan is
$1.221\times10^{-4}$~N. An additional non-collinear state gives a maximum
moment imbalance of $2.052\times10^{-6}$~N\,m after translating moments
to a common origin. Central differences at $0.5~\mu$m translation steps
agree with the two wheel-radial tangent columns within 0.2113\% and
0.2110\% for the reference and coupled cases, below the prescribed 2.5\%
criterion.

The direct reverse-mode comparison exceeds its predefined
$2\times10^{-4}$ scaled-matrix tolerance, with the discrepancy
concentrated in moment rows. A moving force arm contributes a geometric
derivative in addition to the derivative of the applied force, as shown
in Eq.~\eqref{eq:assembled_contact_tangent}. A separate moment assembly
using the unchanged contact forces and sleeve displacement restores
this term and agrees with the production directional tangent at the
tested collinear and non-collinear states. Appendix~\ref{app:moment_diagnostic}
retains the original failed comparison and the complete diagnostic
values. Together with reaction balance and finite differences, this
supports the smooth-state coupling in the scan; contact switching and
other spatial branches require their own assessment.

The example connects the physical load-transfer mechanism to the
off-diagonal tangent blocks required by the assembly solver. Sharing a
pin affects both the reactions and their dependence on the connected
bodies' motion. The structural-port and shared-pin examples identify
these mechanisms under separate prescribed conditions. Their individual
contributions to the following reducer response have not been isolated.

\section{Transmission characteristics of a double-wheel cycloidal reducer}
\label{sec:applications}

\subsection{Assembly and loading protocols}

The assembly example examines how the external torque--angle
response relates to interface loading and local contact pressure in a
double-wheel single-stage cycloidal reducer. Cycloidal gearing,
eccentric bearings, output contacts, and the remaining shaft and main
bearings form the mechanical load path. The assembly geometry, fits,
and profile modifications are held fixed in the calculations but
are not disclosed. The reported results therefore demonstrate the
connection between mechanical observations; they do not specify a
publicly reproducible geometric model of this assembly. The independent
idealized component examples retain their separately declared inputs.

Two protocols use the 24-coordinate model family. A reference-load
state at 140~N\,m supplies contact loads for local pressure recovery.
A signed torque-loading sequence records output deflection and contact
load statistics at 74 levels through loading, unloading, and reversal.
The housing and input rotation are fixed, wheel tilts are constrained,
and the selected input and flange tilt coordinates are retained.
The sequence uses a nominal time step of $2~\mu$s, eight axial
integration slices in the eccentric-bearing contacts, and commanded
ramp and dwell counts of 600 and 6000. Kinetic resets are requested
at each load level and at an interval of 250 steps. These resets are
numerical relaxation operations, distinct from physical damping.
The application concerns these fixed mechanical settings and the
140~N\,m loading range.

\subsection{The internal load path at the reference torque}

\begin{figure}[pos=tbp]
\centering
\includegraphics[width=\linewidth]{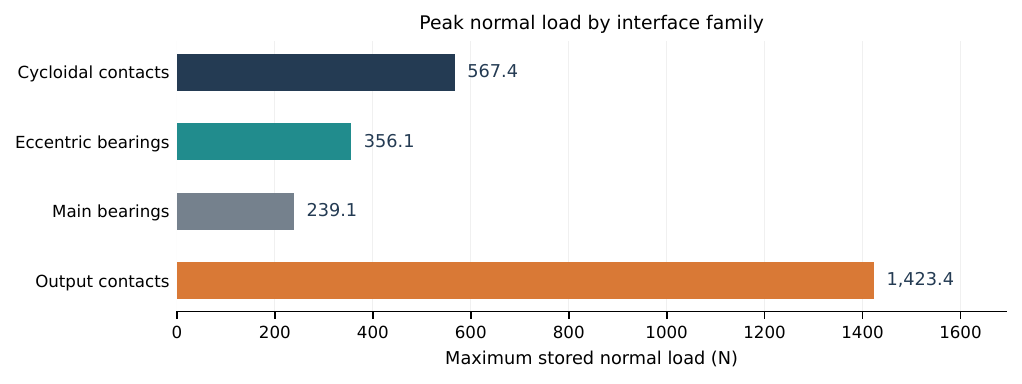}
\caption{Peak normal contact loads in four interface families at
140~N\,m: cycloidal gearing, eccentric bearings, main bearings, and
output contacts. The grouped maxima summarize this operating point
and are not full-cycle load envelopes.}
\label{fig:rated_loads}
\end{figure}

An applied output torque is transmitted through several contact
families. \figref{fig:rated_loads} summarizes the maximum normal
contact loads in cycloidal gearing, eccentric bearings, main bearings,
and output contacts. The two wheel interfaces have peak pin forces of
567.36 and 530.28~N. These maxima identify the largest element loads
within the selected groups.
The transmitted torque itself follows from force directions and
moment arms in Eq.~\eqref{eq:contact_mapping}; a peak normal force
alone is not a transmitted resultant.

The state's kinetic energy is $7.89\times10^{-7}$~J. This provides
a settling diagnostic, but does not establish a force-equilibrium
tolerance. The load summaries are therefore interpreted at that finite
state. Their underlying contact records provide the forces and local
geometry used by the pressure calculation, while the published view
retains only the stated interface-level diagnostics.

\subsection{Loading path and branch-dependent torsional response}

The signed sequence tests whether one torsional stiffness describes
loading, unloading, and reversal. Output torque, output angle, and
contact loads are available at 74 levels. For each branch, the
high-load response is represented by
\begin{equation}
 T=K_{\mathrm{fit}}\theta+b,\qquad
 K_{\mathrm{fit}}=
 \frac{\sum_j(\theta_j-\bar\theta)(T_j-\bar T)}
      {\sum_j(\theta_j-\bar\theta)^2},
 \label{eq:stiffness_fit}
\end{equation}
using points with $70\leq |T|\leq140$~N\,m.
The loading and unloading branches are fitted separately.
The resulting slopes lie between 73.68 and 76.34~N\,m/arcmin
(Table~\ref{tab:branch_stiffness}), with lower fitted slopes on unloading
than on loading. These values describe the stated load interval and
branch; they do not replace the local tangent of the mechanical model.

\begin{table}[pos=tbp]
\caption{Branch fits over $70$--$140$~N\,m in absolute torque.
Negative-branch coordinates are treated with their signed values in the
regression; absolute values are used only for panel comparison.}
\label{tab:branch_stiffness}
\centering
\begin{tabular}{@{}lrrr@{}}
\toprule
Branch & Points & $K_{\mathrm{fit}}$ (N\,m/arcmin) & $R^2$\\
\midrule
Positive loading & 11 & 76.3424 & 0.998003\\
Positive unloading & 9 & 74.3136 & 0.999379\\
Negative loading & 9 & 76.1333 & 0.998788\\
Negative unloading & 9 & 73.6753 & 0.999371\\
\bottomrule
\end{tabular}
\end{table}

\begin{figure}[pos=tbp]
\centering
\includegraphics[width=\linewidth]{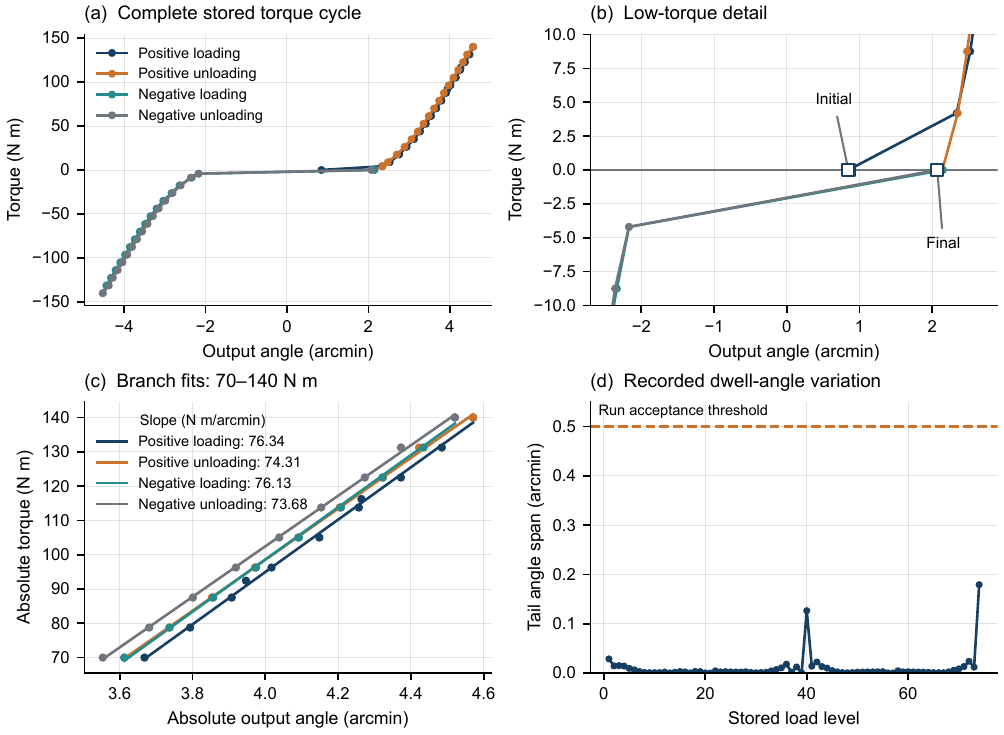}
\caption{Torque-loading path and branch fits for the cycloidal reducer.
(a) Complete signed torque cycle.
(b) Low-torque detail, including distinct initial and final zero-load states.
(c) High-load branch fits using Eq.~\eqref{eq:stiffness_fit}.
(d) Dwell-angle variation at each load level and the protocol's
acceptance threshold. Lines connect the computed points; no smoothing
or closure correction is applied.}
\label{fig:hysteresis}
\end{figure}

The low-torque region departs from these high-load fits. The first zero-load
angle is 0.8462~arcmin, whereas the final zero-load angle is 2.0694~arcmin.
Their difference is 1.2232~arcmin, and the final return to zero torque contains
a pronounced state change. Only one loading cycle is available. The data
therefore establish a branch-dependent response under this relaxation
protocol, but not a stable, repeatable closed hysteresis loop.
Physical history and incomplete stabilization have not been separated;
the enclosed area cannot be assigned to periodic dissipation, and the
initial-to-final offset is not a standardized lost-motion measurement.
The run used no forced commits and its maximum measurement-stage
tail-angle span was 0.1792~arcmin, below the internal threshold of
0.5~arcmin. Passing this local settling criterion did not establish
closure of the complete cycle.

\subsection{Contact-load diagnostics through the loading cycle}

\begin{figure}[pos=tbp]
\centering
\includegraphics[width=\linewidth]{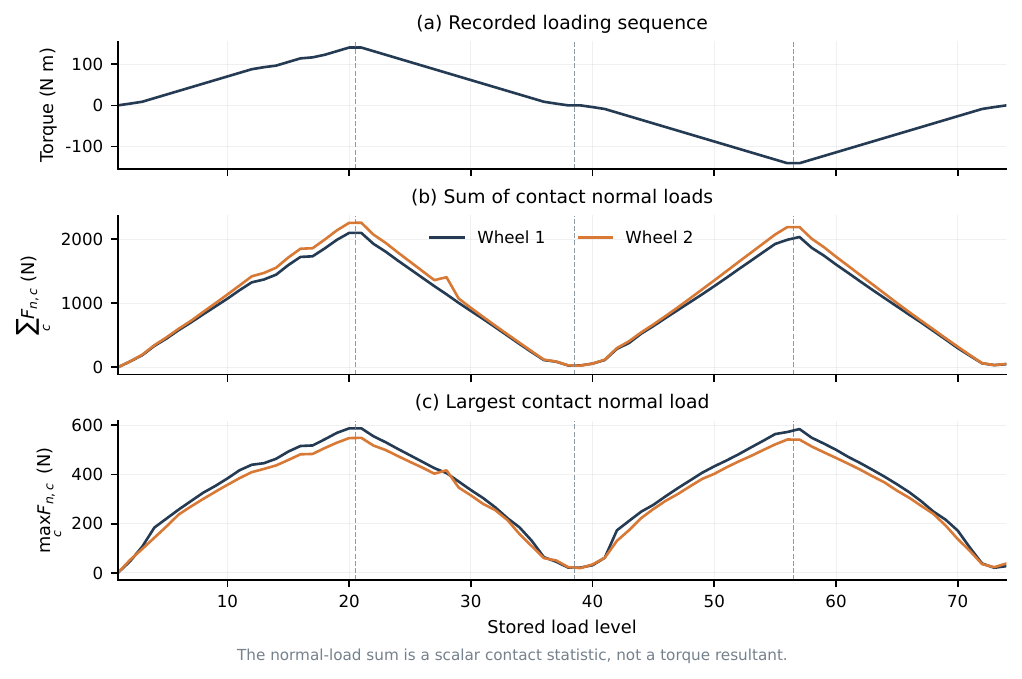}
\caption{Output torque and wheel-contact load diagnostics over the
same 74-level sequence as \figref{fig:hysteresis}. The curves show
the sum and maximum of scalar normal forces. Their sum is not a
resultant. Load-level index
describes the prescribed loading path, not elapsed time or crank angle.}
\label{fig:loading_path}
\end{figure}

The same 74-level sequence permits the external response to be
examined alongside two contact-load diagnostics for each wheel:
the sum of normal-force magnitudes $\sum_c F_{n,c}$ and the largest
element force $\max_c F_{n,c}$. \figref{fig:loading_path} plots these
quantities with output torque against load-level index. The sum measures
aggregate normal loading and the maximum measures the most highly
loaded element; their different variations retain information beyond
the applied torque alone. The sum is a scalar diagnostic, not a
resultant force or torque, because the element directions and lever
arms are not included in it.

Torque, displacement, and these contact statistics belong to the same
mechanical states, so their association does not require a separate
load-sharing model. Isolating the contribution of an individual
structural compliance would require a controlled change under
matched boundaries.

\subsection{From interface loads to local pressure and stress}

\begin{figure}[pos=tbp]
\centering
\includegraphics[width=\linewidth]{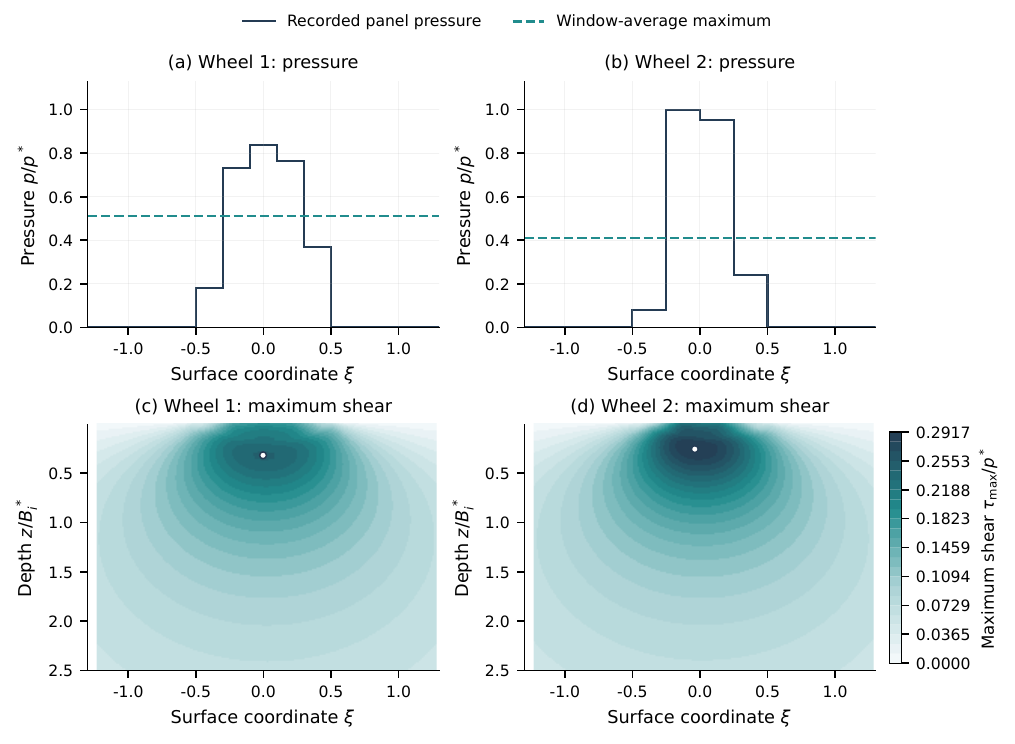}
\caption{Normalized local pressure and maximum-shear fields for
selected contacts at 140~N\,m, using
Eq.~\eqref{eq:public_pressure_normalization}. Pressure retains its
piecewise-constant panel representation, and both shear fields share
the same normalized stress scale. These are normalized presentations
of the original 257-panel pressure and stress fields, not new field
solutions. Section~\ref{sec:pressure_resolution} separately examines
pressure resolution.}
\label{fig:stress}
\end{figure}

Local pressure depends on the geometry carrying a force as well as its
magnitude. The reference-load state is therefore passed to the elastic-contact
postprocessor in Section~\ref{sec:observables}, retaining the selected
contact's pose and profile. \figref{fig:stress} shows one contact per
wheel, selected by the fixed-width pressure measure defined below.
These contacts need not carry the family-wise maximum force summarized
in \figref{fig:rated_loads}.

Two pressure quantities are distinguished:
\begin{equation}
 p_{\mathrm{raw}}=\max_x p(x),\qquad
 \bar p_{w,\max}=\max_s\frac{1}{w}
               \int_{s-w/2}^{s+w/2}p(x)\,\mathrm dx.
 \label{eq:pressure_metrics}
\end{equation}
The physical window $w$ is held fixed between resolutions. Pressure
outside the contact support is zero, and the denominator remains $w$.
This average is a specified spatial measurement functional, not a
substitute for peak pressure. The subsurface calculation uses the
original traction distribution rather than this average.

To present the field shapes without disclosing local geometric or
stress scales, define
\begin{equation}
 \xi=\frac{x-x_c}{B^\ast},\qquad
 \widehat z=\frac{z}{B^\ast},\qquad
 \widehat p=\frac{p}{p^\ast},\qquad
 \widehat\tau=\frac{\tau_{\max}}{p^\ast}.
 \label{eq:public_pressure_normalization}
\end{equation}
Here $\tau_{\max}$ denotes maximum shear stress at each field point.
For each contact, $B^\ast$ is the span of its loaded panels in the
257-panel solution, and $x_c$ is the midpoint of that interval.
The common stress reference $p^\ast$ is the larger of the two
257-panel raw pressure peaks. These reference values are not
disclosed. A contact's spatial normalization remains unchanged
throughout the resolution comparison, and both contacts use the
same pressure scale.

The second selected contact carries a smaller normal force but has
a higher normalized pressure maximum. Thus, the force ranking alone
does not determine the local stress ranking. In the displayed
257-panel solutions the
fixed window covers each complete pressure support, so its averaged
value is controlled by the load integral and can remain unchanged
despite variation in pressure shape. The normalized fields preserve
this distinction without reporting absolute stress or length scales.
Their half-space recovery uses the contact's equivalent material.
The traction interpolation, spectral integration, and stress-grid
resolution require separate checks before a strength margin or
fatigue life can be inferred; normalization does not supply those checks.

\subsection{Pressure resolution at a fixed mechanical state}
\label{sec:pressure_resolution}

The next comparison examines pressure resolution while the mechanical
state, selected contacts, material compliance, and profile representation
remain fixed. The geometry-resolved inputs for the assembly case are
not disclosed. The comparison therefore reports dimensionless fields
and relative changes using Eq.~\eqref{eq:public_pressure_normalization}.
The geometric construction and local solution procedure are described
in Appendix~\ref{app:pressure_reproduction} without releasing the
particular component dimensions.

Before refinement, reconstruction reproduces the original 257-panel
coordinate arrays exactly; maximum pressure differences relative to
the original peaks are below $5\times10^{-15}$. This establishes the
internal correspondence of the fields being compared. Because the
underlying geometry is withheld, this correspondence is not presented
as a publicly reproducible verification of the complete assembly.

The relative difference between two piecewise-constant pressure fields is
\begin{equation}
 e_1(N,M)=\frac{\int |p_N(x)-p_M(x)|\,\mathrm{d}x}
                       {\int p_M(x)\,\mathrm{d}x}.
 \label{eq:pressure_field_difference}
\end{equation}
Integration uses the union of the panel edges, with zero pressure
outside each domain. A common fixed scaling of position and pressure
for each contact leaves this relative measure unchanged.

\begin{table}[pos=tbp]
\caption{Normalized pressure-resolution results at a fixed mechanical
state. The references $p^\ast$ and $B_i^\ast$ are defined in
Eq.~\eqref{eq:public_pressure_normalization}; their dimensional values
are not given. $e_1$ compares each field with the finite 4097-panel
reference. Dashes omit the reference's zero self-difference.}
\label{tab:pressure_resolution}
\centering\small
\setlength{\tabcolsep}{5pt}
\begin{tabular}{@{}rrrrrrr@{}}
\toprule
 & \multicolumn{3}{c}{Wheel 1}
 & \multicolumn{3}{c}{Wheel 2}\\
\cmidrule(lr){2-4}\cmidrule(l){5-7}
$N$ & $p_{\rm raw}/p^\ast$ & $B/B_1^\ast$ & $e_1$ (\%)
    & $p_{\rm raw}/p^\ast$ & $B/B_2^\ast$ & $e_1$ (\%)\\
\midrule
129 & 0.8780 & 1.1953 & 42.50 & 0.8939 & 0.9961 & 45.35 \\
257 & 0.8378 & 1.0000 & 17.79 & 1.0000 & 1.0000 & 28.06 \\
513 & 0.8372 & 0.9018 & 8.51 & 1.0331 & 0.7515 & 12.54 \\
1025 & 0.8390 & 0.9026 & 4.59 & 1.0481 & 0.7522 & 7.56 \\
2049 & 0.8427 & 0.9031 & 2.81 & 1.0824 & 0.7526 & 4.16 \\
4097 & 0.8846 & 0.8907 & -- & 1.0938 & 0.7371 & -- \\
\bottomrule

\end{tabular}
\end{table}

\begin{figure}[pos=tbp]
\centering
\includegraphics[width=\linewidth]{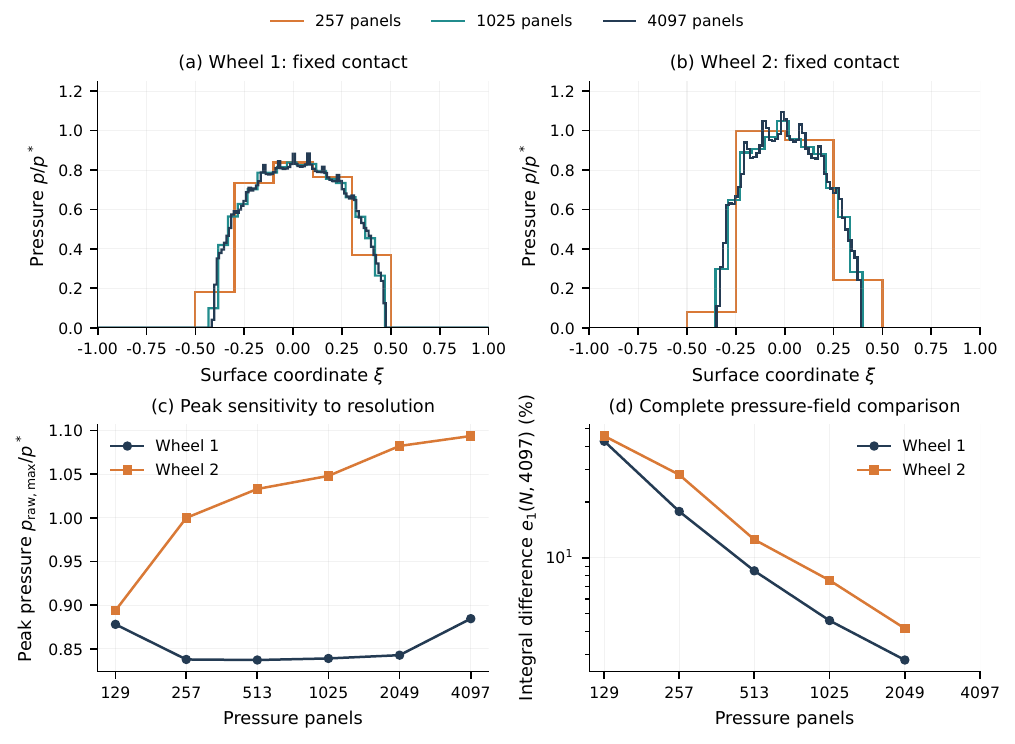}
\caption{Pressure resolution with geometry, load, and state fixed.
(a,b) Dimensionless piecewise-constant pressures at three resolutions.
(c) Normalized raw maxima over all six resolutions.
(d) Integral field differences relative to the finite 4097-panel
reference; its zero self-difference is omitted from the logarithmic plot.
Each contact uses one fixed spatial scale and both use the common
pressure reference throughout the comparison.}
\label{fig:pressure_resolution}
\end{figure}

Scalar stability can conceal remaining field differences. From 513 to
1025 panels, the raw peaks change by 0.22\% and 1.45\%, and both
loaded spans change by only 0.098\%. Nevertheless,
$e_1(513,1025)$ is 8.00\% and 12.88\%.
For a fixed domain length $D$ with contiguous active panels,
$B_N=m_ND/N$, where $m_N$ is the active-panel count. A doubling of
$m_N$ between these nearly doubled resolutions can therefore produce
an almost unchanged span while individual edges and pressures change.

Between 2049 and 4097 panels, the raw peaks change by 4.96\% and
1.05\%, and the spans change by 1.36\% and 2.06\%.
Neither contact satisfies the study's joint successive-change targets
of 2\% for the peak and 1\% for the span. The corresponding integral
field differences are 2.81\% and 4.16\%. At every level the relative
load residual is below $3.6\times10^{-15}$, pressures are nonnegative,
and the loaded region remains inside the computational domain.
Algebraic satisfaction of the discrete equations is thus distinct
from spatial resolution of the pressure field.

The normalized presentation retains this numerical conclusion while
withholding the dimensional contact inputs. The fixed profile polygon
has its own unassessed representation error, and the subsurface recovery
requires separate resolution checks. The assembly example is consequently
used to illustrate the connection between a mechanical state and local
outputs, rather than to establish geometry-specific strength or life.

\section{Modelling implications and scope}
\label{sec:discussion}

\subsection{Preserving interaction when changing the model representation}

The framework's central modelling choice is to retain the relationship
between a contact and the assembly carrying its reactions. Geometry
determines where an interface can transmit force, while the connected
bodies and structure determine its loaded position. The work-conjugate
maps in Eqs.~\eqref{eq:contact_mapping} and
\eqref{eq:port_runtime_transform} preserve this relationship when
local forces are assembled or physical interface components are
represented by fewer structural coordinates. This provides a defined
basis for exchanging a component model without independently prescribing
the loads at each of its interfaces.

The shared-pin result makes the consequence tangible. On the prescribed
radial scan and fixed sleeve-seating branch, displacing one wheel changes
the load on another wheel held at a fixed pose because both load the same
deformable shaft. Removing only the cross-station compliance removes
this incremental radial transfer while preserving each station's
self-compliance. Other directions can remain coupled through the common
sleeve and flange. A load-sharing approximation must therefore be judged
by the interaction it retains, as well as by its local force law.
The corresponding system interpretation is given by
Eq.~\eqref{eq:obs_condensed_stiffness}: output compliance includes the
relaxation of the internal coordinates. Fixing those coordinates changes
the quantity being predicted.

The structural comparison addresses a related choice at a different level.
Retaining the same contact ports does not require retaining the same
number of dynamic coordinates. However, similar static port errors at
nearly equal reduced orders do not imply similar transmission of a
time-varying force. The frequency range carried by the selected modes
matters, as illustrated by the cross-port responses. A static local
correction supplies an additional compliance contribution only where it
is actually connected; it cannot supply omitted dynamic resonances.
These distinctions complement flexible-reducer
studies~\cite{g_li2025_flexible} by making the representation and
interface retained in each comparison explicit.

\subsection{Connecting performance measures through mechanical information}

The assembly example links the output torque response to aggregate
and maximum contact loads from the same loading protocol. Their
different variation along the branches retains information about
internal loading beyond output torque alone. The separate
operating-point protocol supplies identifiable loads and geometry
for local pressure recovery, displayed here in normalized form. These
connections let an external response be interpreted through its internal
load path while preserving the different boundary and observation
conditions of each calculation.

The same principle organizes the broader performance routes in
Table~\ref{tab:performance_protocols}. Precision and stiffness can be
related through a fixed equilibrium branch. Vibration requires the
excitation history and a specified response coordinate. Heat, wear,
and fatigue require different statistics of the contact history and
additional constitutive information. A change in a thermal or wear
field affects subsequent characteristics through its assigned geometry
or friction input and the resulting mechanical redistribution.
This organization makes both the shared mechanical information and the
additional assumptions visible. It also provides a consistent starting
point for a focused study of one characteristic without duplicating the
entire reducer formulation. The detailed geometry and geometry-resolved
application records are withheld; the reported aggregate and normalized
results should therefore be interpreted within that disclosure boundary.

\subsection{Resolution and limits of the present evidence}

There are distinct resolution choices at the structural, contact, and
temporal levels. The annular example assesses a second structural
reduction relative to a fixed parent Craig--Bampton model. The pressure
example holds the body state and polygon fixed while refining local
panels. The oscillator isolates smooth-problem time discretization.
Their error measures concern different operators and cannot be
substituted for one another. In particular, small force residuals
establish solution of a discrete pressure problem, whereas pressure
peaks and individual contact edges can remain sensitive to its
discretization. A fixed-width load average may conceal that sensitivity.

The quantitative demonstrations have corresponding physical boundaries.
The annulus uses an idealized component and prescribed mounting springs;
the shared-pin calculation isolates a local mechanism. The assembly
loading sequence contains one cycle with a nonzero return offset, and
the fixed-snapshot pressure study does not establish converged
material-limit stresses. These are numerical studies rather than
product-calibrated experiments. Repeatable loading cycles, full-period
motion records, and measured assembly responses are needed for
device-specific predictions, consistent with recent experimental
stiffness and hysteresis work~\cite{g_cheng2023_stiffness,g_xu2026_hysteresis}.

GPU batching and eligible captured execution support the computational
organization, but the present results do not supply an end-to-end
speedup benchmark. Such a comparison would require matched mechanical
models, resolution targets, and observation costs on specified hardware.
The contribution established here is the formulation and demonstrated
preservation of load-path information across local calculation,
structural representation, and performance observation.

\section{Conclusions}
\label{sec:conclusion}

A dynamic toolkit for precision reducers has been developed around
explicit contact geometry and a common assembly of motion, reaction,
and deformation. Work-conjugate interface maps connect gear, bearing,
and output contacts to rigid and reduced structural coordinates.
The implicit solution retains a defined distinction between trial
evaluation, accepted history, and the observations passed to performance
protocols and configured slow feedback.

The structural and shared-pin studies demonstrate why these connections
matter. For the annular component, two reductions with similar static
errors have cross-port response errors of 24.975\% and 0.312\% over
the selected frequency band, showing that structural order must be
chosen with the transmitted response in view. In the shared-pin case,
a $20~\mu$m radial displacement at one wheel transfers approximately
75~N to a second wheel held fixed. Removing the cross-station compliance
removes this incremental transfer on the tested sleeve-seating branch.
Together, the studies distinguish local compliance
from the interactions carried by a shared structure.

The double-wheel cycloidal application connects branch-dependent
torsional response to internal contact-load variation. Its fitted
branch slopes span 73.68--76.34~N\,m/arcmin under the stated loading
protocol. The 1.223~arcmin zero-load return difference limits the
interpretation of the single recorded cycle. Local pressure recovery
further demonstrates the reuse of a specified mechanical state,
while the panel study exposes pressure and contact-edge sensitivity
that a load average does not reveal.

The numerical demonstrations concern mechanical coupling, loading response,
and pressure recovery. The further performance and slow-state formulations
define how the same motion and contact records are consumed by precision,
vibration, thermal, wear, and durability models. Component verification
supports the selected contact arithmetic and smooth-problem integration
behaviour; the present studies do not establish complete reducer accuracy,
repeatable hysteresis, material-limit stresses, or computational speedup.
The reusable result is a
formulation in which geometry changes, load redistribution, and
external response can be followed through the same physical interfaces.

\FloatBarrier
\section*{Acknowledgments}
This work was supported by the Jiangsu Funding Program for Excellent
Postdoctoral Talent (Grant No.~2025ZB195). The authors acknowledge the
open-source multibody and GPU computing projects cited in this work.
\section*{Data availability}
The appendices in this document specify the extended formulations and
the parameters of the independent idealized component examples.
The public source package contains the manuscript and the figure assets
required to compile it. Detailed geometry, fits, modification values,
dimensional pressure references, and geometry-resolved records for the
reducer application are not disclosed. That application is presented
through aggregate load statistics and normalized local fields; it is
not a publicly reproducible geometric model of the complete reducer.
The original calculation library and raw run archives are not included
in the public source package.
\section*{Declaration of competing interest}
The authors declare no competing interests.
\section*{CRediT authorship contribution statement}
\textbf{Jiacheng Miao:} Writing -- original draft, Writing -- review \& editing,
Software, Validation, Visualization, Data curation, Conceptualization.
\textbf{Chao Liu:} Project administration, Supervision, Resources, Validation.
\textbf{Qiliang Wang:} Supervision, Project administration, Resources, Funding acquisition.
\textbf{Yunhui Guan:} Supervision, Writing -- review \& editing, Methodology.
\textbf{Weidong He:} Supervision, Writing -- review \& editing, Methodology.
\appendix
\section{Slow-process models and feedback protocols}
\label{sec:supp_slow_processes}

The main article connects accepted mechanical histories to performance
observations and to configured changes of contact geometry or friction.
This section supplies the corresponding slow-process formulations. They
describe implemented feedback paths and the additional constitutive inputs
required to use them; the mechanical and pressure examples in the main
article do not provide thermal or durability calibration.

For self-contained notation, let $F_{c,n}$ and $v_{s,c,n}$ be the normal
force and slip speed of contact element $c$ at accepted evaluation $n$,
with actual accepted duration $\Delta t_n^a$. A representative mechanical
window of duration $\tau_w$ provides
\begin{equation}
 I_{s,c}=\sum_n F_{c,n}|v_{s,c,n}|\Delta t_n^a,
 \qquad
 I_{m,c}=\sum_n F_{c,n}^{m}\Delta t_n^a .
 \label{eq:supp_slow_exposure}
\end{equation}
The exponent $m$ belongs to the consuming life law. Only accepted durations
contribute to these exposures; repeated nonlinear evaluations do not add
physical time. Each consumer specifies a window covering the relevant
mechanical or material-frame cycle. The mechanical window, thermal update
interval $\Delta\tau_T$, and represented service interval
$\Delta\tau_{\mathrm{service}}$ are distinct time scales.

\subsection{Thermal and lubrication feedback through geometry and friction}

The sliding statistic $I_s$ is a common input to heat and wear, but is not
itself frictional heat. For a declared sliding-loss model, a window-mean
source is $\overline P_c=\mu_c I_{s,c}/\tau_w$. A tangential law with
recoverable elastic storage instead requires a compatible dissipation
definition. The thermal model assigns sources to physical nodes and solves
\begin{equation}
 \boldsymbol{C}_T\dot{\boldsymbol{T}}
 +\boldsymbol{L}_T\boldsymbol{T}=\overline{\boldsymbol{Q}},
 \qquad
 \left(\frac{\boldsymbol{C}_T}{\Delta\tau_T}
 +\boldsymbol{L}_T\right)\boldsymbol{T}^{k+1}
 =\frac{\boldsymbol{C}_T}{\Delta\tau_T}\boldsymbol{T}^{k}
 +\overline{\boldsymbol{Q}}^{k}.
 \label{eq:obs_thermal}
\end{equation}
Here $\boldsymbol T$ contains the nodal temperatures, $\boldsymbol C_T$
collects heat capacities, $\boldsymbol L_T$ is the thermal conductance
operator, and $\overline{\boldsymbol Q}$ is the assigned mean source
vector. Heat capacities, conductances, source partitions, and ambient
conditions are model inputs. Contact-dependent conductances can be updated at thermal
boundaries. For steady operation, the alternative protocol iterates the
mechanical source window and thermal equilibrium with under-relaxation;
both the unrelaxed temperature residual and the source-vector drift are
checked. This iteration seeks a thermal fixed point; its iteration count is
not elapsed warm-up time. Several mechanical phase samples can share one
candidate temperature field, so their weighted sources are combined before
solving the common thermal problem.

A minimal instance retains one contact family's heat source, one
internal thermal node, and a conductance $G$ to an ambient node. Its update
reduces to
\begin{equation}
 T^{k+1}=
 \frac{(C_T/\Delta\tau_T)T^k+
       \mu\sum_c I_{s,c}/\tau_w+GT_{\mathrm{amb}}}
      {C_T/\Delta\tau_T+G}.
 \label{eq:obs_single_node}
\end{equation}
Writing the resulting thermal gap and reevaluating the same mechanical
state isolates the feedback effect; subsequently advancing the assembly
reveals redistribution through the other contacts. This separates source
accumulation, the thermal update, the geometry map, and the mechanical
consequence into reproducible parts.

Temperature changes the mechanics only through configured feedback maps.
For example, a bearing clearance increment contains the outer-race
expansion minus the inner-race and rolling-element expansions. The cycloid
radial-envelope map is
\begin{equation}
 \Delta g_T=
 \alpha_h\Delta T_hR_z-\alpha_p\Delta T_pr_z
 -\alpha_w\Delta T_w(R_z-r_z).
 \label{eq:obs_thermal_gap}
\end{equation}
Here $R_z$ is the pin-centre-circle radius, $r_z$ is the reference pin
radius supplied to this envelope map, and
the subscripts $h$, $p$, and $w$ identify the housing, pins, and wheel.
The coefficients $\alpha$ describe linear expansion, and each $\Delta T$
is measured from the corresponding reference temperature.
The radius $R_z-r_z$ is the effective wheel radius used by this envelope
approximation.
Equal expansion coefficients and equal temperature increments cancel to
first order in Eq.~\eqref{eq:obs_thermal_gap}; differential expansion
determines the gap change.
Uniform expansion, support-center motion, and local thermal shape have
separate channels; external finite-element displacements can be projected
onto the supplied contact normals. These offsets enter the next contact
evaluation and can change the loaded set, preload, stiffness, and precision.

The lubrication extension evaluates viscosity $\eta(T)$, film thickness,
and $\lambda=h_{\min}/\sqrt{R_{q1}^2+R_{q2}^2}$, then derives a mixed
friction coefficient; $R_{q1}$ and $R_{q2}$ are the RMS surface roughnesses.
In observation mode these values are outputs only.
In coupled mode the coefficient is written for the next mechanical window.
The present implementation applies a contact-instance coefficient for
cycloid meshes and supports element-wise coefficients for the involute
path. This is a staggered constitutive update, not a monolithic thermal-EHL
solution. Its heat-source estimate is not automatically identical to the
mechanical friction work in the preceding window. Temperature-dependent
modulus in the film calculation likewise does not establish a general
temperature-dependent structural stiffness update.

\subsection{Material evolution, damage, and renewed mechanical observations}

For a declared nominal contact area $A_c$, the wear update has the form
\begin{equation}
 \Delta h_c=k_w(\lambda_c)\frac{I_{s,c}}{A_c}\,s,
 \qquad s=\frac{\Delta\tau_{\mathrm{service}}}{\tau_w}.
 \label{eq:obs_wear}
\end{equation}
The coefficient $k_w$ is the declared wear-law input at the supplied film
parameter $\lambda_c$; its selection is separate from the mechanical load
calculation.
The material-loss increment is written to the wear contribution of the
contact gap. The single-channel integrator limits the largest increment
and reduces the represented service time by the same factor. Subsequent
mechanical settling is excluded from the next extrapolation window. This
is essential: a geometry-update transient is not a stationary load spectrum
that can be repeated for many service hours. The association between contact
elements and material surfaces also determines the spatial meaning of the
update: a tooth-averaged gap increment does not resolve a local wear scar.
In a multiple-channel workflow, independently limited updates can represent
different service times; a common admissible
increment is therefore an additional composition requirement. Temperature-
or film-dependent wear coefficients likewise require an explicit refresh
schedule, beyond their availability as constitutive functions.

For a calibrated force-life relation
$N(F)=N_{\mathrm{ref}}(F_{\mathrm{ref}}/F)^m$, the damage increment is
\begin{equation}
 \Delta D_c=
 \frac{\nu_c\Delta\tau_{\mathrm{service}}}
      {N_{\mathrm{ref}}F_{\mathrm{ref}}^m}
 \frac{I_{m,c}}{\tau_w},
 \label{eq:obs_damage}
\end{equation}
Here $N_{\mathrm{ref}}$ is the calibrated life in cycles at reference force
$F_{\mathrm{ref}}$, and $\nu_c$ is the declared stress-cycle rate. For $m>1$, replacing the
load moment by the power of the mean load underestimates the contribution
of load variation. Separate
stress-spectrum calculations retain $D=\sum_r n_r/N(\sigma_r)$, loaded
flank, and cycle counts. Bearing rating and material fatigue therefore
require their own calibrated inputs; a stored load maximum is insufficient.
Damage is a downstream state unless a spatial defect adapter is explicitly
connected. A scalar clearance change cannot represent the position and
material-frame motion of a localized race defect.

After a slow-state update, the precision, stiffness, and vibration protocols
can be repeated at the new state with matched loading and installation
conditions. The resulting changes then have an identifiable origin:
modified contact geometry or friction, altered load distribution, and a
different assembly response. The implemented thermal, lubrication, wear,
and fatigue protocols use these shared physical channels with their own
update schedules. Combining them requires compatible state updates and
time accounting; their individual availability does not establish a
general simultaneous multiprocess calculation.

\section{Reproducing the shared-pin calculation}
\label{app:reproduction}

\subsection{Inputs, evaluation sequence, and saved quantities}

The main-text study entitled ``Load transfer through a shared output
pin'' requires only the two wheel poses, flange pose, and the parameters
repeated in Table~\ref{tab:supp_pin_inputs}.
It does not require constructing a complete reducer or generating a shaft
finite-element model. The recorded calculation uses the production
output-pin force and directional-tangent functions. The inputs and
evaluation sequence below define the prescribed local problem, including
its coordinate order and numerical regularization. Exact replay of the
implementation requires the matching calculation code and original
records, which are not included in the public manuscript source package.

The nine-coordinate order is
\begin{equation}
 \boldsymbol q=(x_f,y_f,\theta_f,x_0,y_0,\theta_0,
                         x_1,y_1,\theta_1)^{\mathsf T},
 \label{eq:supp_pin_coordinates}
\end{equation}
where $f$ denotes the flange. Translations use metres and rotations
use radians about the $z$ axis; all axial translations and other
rotations are constrained. The reaction triples are $(F_x,F_y,M_z)$
about the corresponding body origins.

\begin{table}[pos=tbp]
\caption{Inputs to the shared-pin reproduction. These repeat the
main-text component inputs so that the local calculation is specified
independently of the complete reducer.}
\label{tab:supp_pin_inputs}
\centering\small
\begin{tabularx}{\linewidth}{@{}p{.43\linewidth}Y@{}}
\toprule
Quantity & Value or convention\\
\midrule
Pin and hole distribution radii & $R=35$~mm for both\\
Outer pin and wheel-hole radii & $1.50$ and $1.55$~mm\\
Whole-contact power-law coefficient & $k=2\times10^8~\mathrm{N}\,\mathrm{m}^{-10/9}$\\
Contact width and integration & $6$~mm; four equal-weight midpoint slices\\
Slice stations relative to each wheel & $\{-2.25,-0.75,0.75,2.25\}$~mm\\
Wheel midplanes relative to flange & $z_0=-3$~mm, $z_1=3$~mm\\
Self- and cross-compliance & $C_{00}=C_{11}=1.5\times10^{-9}$~m/N;
 $C_{01}=0$ or $0.9\times10^{-9}$~m/N\\
Sleeve model & Massless; $5~\mu$m radial clearance disk;
 inner Hertz indentation omitted\\
Prescribed translations & $x_0=100$--$120~\mu$m at 81 equally spaced
 points; $x_1=-70~\mu$m; reference $x_0=110~\mu$m\\
Remaining inputs & One pin; zero phases, other coordinates, velocities,
 damping, and thermal offsets; no tangential friction\\
Numerical regularization & Sleeve-update tangent floor $10^3$~N/m;
 determinant floor $10^{-12}$; distance and open-tangent indentation
 floors $10^{-9}$~m\\
\bottomrule
\end{tabularx}
\end{table}

For wheel $i$ and slice $s$, let $d_{is}$ be the geometric approach
at the current sleeve position and let $\boldsymbol n_i$ point in the
force direction on that wheel. The local calculation uses
\begin{equation}
 \begin{aligned}
 F_i^0&=\sum_s(k/4)[d_{is}]_+^{10/9},&
 K_i&=\sum_{s:d_{is}>0}\frac{10k}{36}
                  \max(d_{is},10^{-9}\,\mathrm m)^{1/9},\\
 \boldsymbol C_p&=
 \begin{bmatrix}C_{00}&C_{01}\boldsymbol n_0^{\mathsf T}\boldsymbol n_1\\
 C_{01}\boldsymbol n_0^{\mathsf T}\boldsymbol n_1&C_{11}\end{bmatrix},&
 \boldsymbol K_c&=\operatorname{diag}(K_0,K_1),\\
 (\boldsymbol I+\boldsymbol K_c\boldsymbol C_p)\boldsymbol F^c
 &=\boldsymbol F^0,&
 \boldsymbol u_p&=\boldsymbol C_p\boldsymbol F^c.
 \end{aligned}
 \label{eq:supp_pin_closure}
\end{equation}
Tensile corrected resultants are clipped to zero. The station deflections
$u_{p,i}$ are then subtracted from $d_{is}$ before nonlinear slice-force
reevaluation. Starting at $\boldsymbol u_{\rm fit}^0=\boldsymbol0$, the
massless sleeve is updated by
\begin{equation}
 \boldsymbol u_{\rm fit}^{k+1}=
 \Pi_{\|\boldsymbol u\|\leq c}
 \left(\boldsymbol u_{\rm fit}^k
       -0.7\frac{\sum_i F_i^c\boldsymbol n_i}
                    {\max(K_0+K_1,10^3~\mathrm{N/m})}\right),
 \qquad c=5~\mu\mathrm m.
 \label{eq:supp_pin_fit}
\end{equation}
Projection retains an interior trial or radially rescales an exterior
trial to radius $c$. The prescribed 16 updates and subsequent nonlinear
reevaluation define a finite-iteration algorithm, not a fully converged
nonlinear complementarity solution.

For the solver-sign tangent $\boldsymbol K_q=-\partial\boldsymbol Q/
\partial\boldsymbol q$, define
\begin{equation}
 \boldsymbol S=\operatorname{diag}(1,1,1/R,1,1,1/R,1,1,1/R),\qquad
 \widehat{\boldsymbol q}=\boldsymbol S^{-1}\boldsymbol q,\quad
 \widehat{\boldsymbol Q}=\boldsymbol S\boldsymbol Q,\quad
 \widehat{\boldsymbol K}=\boldsymbol S\boldsymbol K_q\boldsymbol S.
 \label{eq:supp_pin_scaling}
\end{equation}
Thus each coordinate triple is $(x,y,R\theta)$, each force triple is
$(F_x,F_y,M_z/R)$, and every tangent entry has units N/m.

For each value of $C_{01}$ and imposed $x_0$, the reproducible sequence is:
\begin{enumerate}
 \item Set $\boldsymbol q$ in the order of
 Eq.~\eqref{eq:supp_pin_coordinates}, set all velocities to zero,
 and initialize the sleeve offset at the centre of its clearance disk.
 \item Evaluate both hole geometries, the four slice approaches per wheel,
 and the open resultants and tangents in Eq.~\eqref{eq:supp_pin_closure}.
 Form the normal-projected compliance and solve
 Eq.~\eqref{eq:supp_pin_closure}.
 \item Apply Eq.~\eqref{eq:supp_pin_fit} and repeat the local
 evaluation for the prescribed 16 updates. At the final sleeve position,
 recompute the local closure and reevaluate the nonlinear slice forces.
 \item Assemble the three body-reaction triples and save both wheel
 resultants, all eight slice forces and approaches, and the sleeve offset.
 Check total force and moments translated to a common origin.
 \item At the reference state, evaluate the production $9\times9$ tangent.
 Use Eq.~\eqref{eq:supp_pin_scaling} before comparing its blocks
 or reporting a matrix norm.
\end{enumerate}
The original two 81-point scans and the additional derivative diagnostic
are separate calculations. The initial failed direct reverse-mode
comparison is retained in the diagnostic table below; it is not
replaced by the later moment-assembly result.
The recorded runs use Warp~1.16.0. Both compliance cases are evaluated
on the same physical GPU. An implementation-level repetition should
likewise record the interpreter, library version, selected physical
adapter, and kernel-build configuration, and use a consistent device
for the direct comparison.

\subsection{Independent check of the geometric moment derivative}
\label{app:moment_diagnostic}

A point force contributes the moment
$M_z=r_xF_y-r_yF_x$ about a body's reference origin. Differentiation requires
both force and geometry terms:
\begin{equation}
 \frac{\partial M_z}{\partial q_j}
 =r_x\frac{\partial F_y}{\partial q_j}
  -r_y\frac{\partial F_x}{\partial q_j}
  +\frac{\partial r_x}{\partial q_j}F_y
  -\frac{\partial r_y}{\partial q_j}F_x.
 \label{eq:moment_chain_rule}
\end{equation}
For the wheel arm $\boldsymbol r=R(\cos\theta,\sin\theta)$,
its purely geometric contribution is
$-R(\sin\theta F_y+\cos\theta F_x)$, reducing to $-RF_x$ at
the collinear reference state. Force derivatives alone therefore do
not supply the complete moment tangent.

The initial reverse-mode comparison differed from the production
directional tangent mainly in the moment rows. To test this contribution
without changing the contact law, the production kernel supplied its planar
forces and sleeve offset to a separate moment-assembly kernel. The
reconstructed generalized force is identical to the production output
at the collinear reference state. At the non-collinear control state,
only the flange moment differs, by less than
$5.74\times10^{-7}$~N\,m owing to the force-summation order.

The additional control retains the same geometry and compliance. The flange
translation is zero, the wheel translations are
$(x_0,y_0)=(70,30)~\mu$m and $(x_1,y_1)=(-60,-25)~\mu$m, and
$(\theta_f,\theta_0,\theta_1)=(10^{-4},2\times10^{-4},-1.5\times10^{-4})$~rad.
All velocities remain zero. It supplements, rather than replaces, the
collinear reference state.

For each comparison method, the reported relative error is
\begin{equation}
 e_K=
 \frac{\left\|\boldsymbol S(\boldsymbol K_{\mathrm{prod}}
                         -\boldsymbol K_{\mathrm{ref}})\boldsymbol S\right\|_{\mathrm F}}
      {\left\|\boldsymbol S\boldsymbol K_{\mathrm{ref}}\boldsymbol S\right\|_{\mathrm F}},
 \label{eq:moment_comparison_error}
\end{equation}
where $\boldsymbol K_{\mathrm{ref}}$ is the reverse-mode matrix identified
by that table column. The denominator is its own scaled norm.

\begin{table}[pos=tbp]
\caption{Relative Frobenius differences defined by
Eq.~\eqref{eq:moment_comparison_error}, after work-conjugate scaling.
Both reverse-mode calculations use unchanged production contact forces.}
\label{tab:moment_diagnostic}
\centering\small
\begin{tabular}{@{}llrr@{}}
\toprule
State & Cross-compliance & Direct reverse mode & Separate moment assembly\\
\midrule
Collinear & $C_{01}=0$ & $4.589\times10^{-4}$ & $8.330\times10^{-8}$\\
Collinear & $C_{01}>0$ & $4.741\times10^{-4}$ & $1.488\times10^{-7}$\\
Non-collinear & $C_{01}=0$ & $5.582\times10^{-4}$ & $1.020\times10^{-6}$\\
Non-collinear & $C_{01}>0$ & $5.561\times10^{-4}$ & $6.148\times10^{-7}$\\
\bottomrule
\end{tabular}
\end{table}

Inspection of the generated reverse code for the direct path showed that
the force-arm adjoint used zero-initialized force accumulators before
replaying the dynamic slice accumulation. This accounts for the missing
geometric term in this test. Independent central differences also support
the rotational columns of the production tangent.
Very small translational perturbations remain limited by single-precision
coordinate subtraction: near a 35~mm coordinate, the spacing is about
3.73~nm. Reducing a perturbation consequently need not improve its
finite-difference estimate.

This diagnostic supports the production tangent at the tested smooth
states and identifies a limitation of the direct reverse-mode path used
here. It is not a general result about automatic differentiation, a
double-precision analytical proof of every matrix entry, or validation
of contact-switching and spatial branches. Both the original failed
comparison and the additional diagnostic are reported here, together
with the conditions under which they were obtained.

\section{Reproducing the structural port comparison}
\label{app:ports_reproduction}

The structural calculation uses the annulus inputs repeated in
Table~\ref{tab:supp_ports_inputs}, corresponding
to the main-text study ``Structural reduction at distributed contact
ports''. The recorded study calls the
production finite-element constructor, Craig--Bampton reduction, and
port reduction functions. The reference arrays are the fixed parent
mass, stiffness, damping, and boundary groups stored with the mesh.
The 176-coordinate model is the parent of the second reduction;
it is not treated as a converged continuum solution.

\begin{table}[pos=tbp]
\caption{Annular component inputs for the structural reproduction.}
\label{tab:supp_ports_inputs}
\centering\small
\begin{tabularx}{\linewidth}{@{}p{.39\linewidth}Y@{}}
\toprule
Quantity & Value\\
\midrule
Inner / outer radius; axial height & 105 / 125 mm; 67.6 mm\\
Young's modulus / Poisson ratio / density & 210 GPa / 0.3 / 7850 kg/m$^3$\\
Contact locations & 52 nodal ports, each with three translations\\
Mounting & 18 nodal spring locations in a lower-face neighbourhood;
 nominal radius 116.5 mm\\
Mount stiffness & $5.556\times10^6$ N/m per location and Cartesian direction\\
Finite elements & First-order displacement tetrahedra; nominal mesh size 6.667 mm\\
Generated mesh & 3638 vertices, 11864 elements, 10914 displacement DOFs\\
Parent Craig--Bampton model & 156 boundary components and 20 fixed-interface modes\\
Damping & Rayleigh calibration with $\zeta=0.02$ at the constructor's
 calibration frequencies; no added mount damping\\
Frequency-response sampling & 10--5000 Hz, increment 2.5 Hz\\
\bottomrule
\end{tabularx}
\end{table}

The original parent matrices, mesh, and port groups define the actual
nodal locations and damping used for the comparison. A repetition of
the prescribed continuum problem can generate a different mesh;
exact reproduction of the reported second-reduction comparisons
requires those same parent matrices. The generated mesh and first
Craig--Bampton reduction are not independently refined here.

The frequency-response input vector has unit global-$x$ component at
port~0. For the cross-port output, only the three entries at port~8 are
nonzero, with direction
$[\cos(16\pi/52),\sin(16\pi/52),0]^{\mathsf T}$.
The comparison samples the complex response over 10--5000 Hz,
including the separate frequency-independent local-residual reconstruction.
Static errors use the reference port compliance
$\boldsymbol C^{\rm ref}=\boldsymbol S_b\boldsymbol K_r^{-1}
\boldsymbol S_b^{\mathsf T}$ and
\begin{equation}
 \epsilon_C=\max_g
 \frac{\|(\boldsymbol C-\boldsymbol C^{\rm ref})_{g,:}\|_{\rm F}}
      {\|\boldsymbol C^{\rm ref}_{g,g}\|_{\rm F}},
 \label{eq:supp_ports_static_error}
\end{equation}
where $g$ selects the three Cartesian rows of one port. The numerator
uses its full $3\times156$ row block and the denominator the corresponding
$3\times3$ self-block. The same denominator is used for every variant.

The additional algebraic map check uses 12 structural modes and two
auxiliary common translations. For the nominal radial direction
$\boldsymbol n_j=[\cos\theta_j,\sin\theta_j,0]^{\mathsf T}$,
$\theta_j=2\pi j/52$, each linear trace measures the difference between
common translation and the corresponding port displacement.
The trace stiffness is prescribed as
$k_j=10^8[1+0.2\cos(2\theta_j)]$~N/m.
Assembling these trace outer products before and after the constant
expanded-coordinate map checks all mixed terms in
\begin{equation}
 \boldsymbol q_e=\boldsymbol T_e\boldsymbol q_s,\qquad
 \boldsymbol Q_s=\boldsymbol T_e^{\mathsf T}\boldsymbol Q_e,\qquad
 \boldsymbol J_s=\boldsymbol T_e^{\mathsf T}\boldsymbol J_e\boldsymbol T_e.
 \label{eq:supp_port_transform}
\end{equation}
Here $\boldsymbol T_e$ retains the integrated coordinates and appends
the physical port displacements through their linear traces.
The trace vectors, map, load and perturbation vectors, and assembled
matrices define the original check. This is a linear algebra test using the generated
structural basis, not a loaded nonlinear gear-contact calculation.

The parent model is constructed once and reused across the reduction
variants. Regenerating it after a change in geometry, meshing,
mounting, or implementation defines a new comparison reference.
The recorded study uses NGSolve~6.2.2606, NumPy~2.4.6, and
SciPy~1.17.1 for offline CPU calculation. The numerical records are
distinct from the GPU shared-pin experiment. The public source package
contains this manuscript and its figure assets; it does not distribute
the original matrices, calculation scripts, or calculation library.

\section{Local pressure reconstruction and comparison procedure}
\label{app:pressure_reproduction}

The pressure example holds a previously computed mechanical state and
its selected interface loads fixed. The component geometry, fits,
profile modification values, and dimensional normalization references
for that assembly are not disclosed. This appendix specifies the
calculation's mathematical inputs and sequence; it does not provide
the geometry-resolved records needed to repeat that particular
assembly calculation. The independently specified component examples
in Appendices~\ref{app:reproduction} and~\ref{app:ports_reproduction}
retain their idealized parameters.

For an arbitrary specified contact, the inputs are a final wheel
profile, mating pin radius, relative pose, selected contact location,
normal force, face width, and the two bodies' elastic constants.
The profile representation remains fixed during a panel-resolution
study. Let $\boldsymbol c$ be the contact point and $\boldsymbol n$
the unit normal in the force direction on the wheel, $r_p$ the mating pin radius, and
$(\boldsymbol u,\theta)$ the wheel pose. The pin centre in wheel
coordinates follows from
\begin{equation}
 \boldsymbol p=\boldsymbol R(\theta)^{\mathsf T}
       (\boldsymbol c-r_p\boldsymbol n-\boldsymbol u).
 \label{eq:public_pressure_pose}
\end{equation}
The selected profile neighbourhood defines a tangent and normal frame.
For cyclic polygon vertices $\boldsymbol r_k$, the production tangent
uses the normalized five-point direction
\begin{equation}
 \boldsymbol d_k=(\boldsymbol r_{k-2}-8\boldsymbol r_{k-1}
                   +8\boldsymbol r_{k+1}-\boldsymbol r_{k+2})/12,
 \qquad \boldsymbol t_k=\boldsymbol d_k/\|\boldsymbol d_k\|.
 \label{eq:public_pressure_tangent}
\end{equation}
Define $\boldsymbol n_k=(-t_{k,y},t_{k,x})$ and let $(p_t,p_n)$
be the projections of $\boldsymbol p-\boldsymbol r_k$ onto
$(\boldsymbol t_k,\boldsymbol n_k)$. Projecting the nearby vertices
into this frame and interpolating them
defines $y_w(x)$. The chosen mating-circle branch is
$y_p(x)=p_n+\sigma\sqrt{r_p^2-(x-p_t)^2}$, with
$\sigma\in\{-1,1\}$ determined by the contact side. A constant datum
is subtracted from $\sigma[y_w(x)-y_p(x)]$ to define the local gap.
That constant shifts the approach reference without changing the
prescribed-force pressure problem.

For each resolution, uniform panel edges cover the same admissible
physical domain, and the gap is sampled at the panel midpoints.
The two-body compliance in Eq.~\eqref{eq:obs_panel_influence} is
unchanged. The local solve enforces nonnegative panel force, a
nonnegative remaining gap, complementarity, and the prescribed
total force. Changing resolution must not reselect a different
governing contact or replace its applied load.

The field comparison in Eq.~\eqref{eq:pressure_field_difference}
is integrated exactly for piecewise-constant pressures by subdividing
on the sorted union of the two sets of panel edges. If
$x=x_c+B^\ast\xi$ and $p=p^\ast\widehat p$, the factor
$B^\ast p^\ast$ cancels between numerator and denominator.
Peak and span ratios likewise use the same reference throughout
each sequence. This permits reporting relative resolution dependence
without publishing its dimensional normalization constants.

The reported sequence does not meet both prescribed peak and span
change targets at its finest pair of resolutions. It therefore
does not establish converged pressure maxima. The existing subsurface
fields are normalized for presentation and are not recomputed by
this panel study. Their traction interpolation, spectral integration,
and observation-grid resolution remain separate numerical questions.

\FloatBarrier

\end{document}